\documentclass[11pt]{article}

\usepackage[margin=1in]{geometry}
\usepackage{setspace}

\usepackage{mathptmx}            
\usepackage[T1]{fontenc}
\usepackage{amsthm,amsmath,amsfonts,amssymb}
\usepackage[authoryear,round]{natbib}
\usepackage{graphicx}
\usepackage{booktabs}
\usepackage{bm}
\usepackage{algorithm}
\usepackage{algpseudocode}
\usepackage{fancyhdr}
\usepackage[colorlinks,citecolor=blue,linkcolor=blue,urlcolor=blue]{hyperref}
\theoremstyle{plain}

\theoremstyle{definition}

\newtheorem*{proofsketch}{Proof sketch}

\newcommand{\R}{\mathbb{R}}
\newcommand{\E}{\mathbb{E}}
\newcommand{\bx}{\mathbf{x}}
\newcommand{\by}{\mathbf{y}}
\newcommand{\bff}{\mathbf{f}}
\newcommand{\btheta}{\bm{\theta}}

\newcommand{\balpha}{\bm{\alpha}}
\newcommand{\bgamma}{\bm{\gamma}}
\newcommand{\GP}{\mathcal{GP}}
\newcommand{\Normal}{\mathcal{N}}
\newcommand{\Kff}{K}
\newcommand{\Ky}{K_{\by}}
\newcommand{\tr}{\operatorname{tr}}
\newcommand{\diag}{\operatorname{diag}}
\DeclareMathOperator*{\argmax}{arg\,max}

\usepackage{xcolor}
\definecolor{tfblue}{RGB}{0,51,141}
\usepackage{titlesec}
\titleformat{\section}{\large\bfseries\color{tfblue}}{\thesection}{0.6em}{}
\titleformat{\subsection}{\normalsize\bfseries\color{tfblue}}{\thesubsection}{0.6em}{}
\titleformat{\subsubsection}{\normalsize\bfseries\itshape\color{tfblue}}{\thesubsubsection}{0.6em}{}
\titleformat{\paragraph}[runin]{\normalfont\bfseries}{}{0pt}{}
\titlespacing*{\section}{0pt}{1.6ex plus 0.4ex minus 0.2ex}{0.8ex plus 0.2ex}
\titlespacing*{\subsection}{0pt}{1.3ex plus 0.3ex minus 0.2ex}{0.6ex plus 0.2ex}
\titlespacing*{\subsubsection}{0pt}{1.1ex plus 0.3ex minus 0.2ex}{0.5ex plus 0.2ex}
\titlespacing*{\paragraph}{0pt}{1.0ex plus 0.2ex}{0.6em}

\makeatletter
\renewcommand{\@maketitle}{%
  \null\vskip 0.3em
  \begin{flushleft}%
    {\large\bfseries\color{tfblue}\@title\par}%
    \vskip 0.7em%
    {\@author\par}%
  \end{flushleft}%
  \vskip 0.4em}
\makeatother

\begin{document}

\title{Sparsity by Default: The Theory and Practice of ARD in
Gaussian Process Regression for Variable Selection}

\author{ Jia Cai \thanks{\textbf{\color{tfblue}CONTACT} : Jia Cai.
Email: jcai8@gmu.edu. Department of Statistics, George Mason University, Fairfax, 
Virginia/USA.}\\[3pt]
{\small jcai8@gmu.edu}\\
{\small Department of Statistics, George Mason University}}

\date{}

\maketitle
\thispagestyle{fancy}

\begin{abstract}
Automatic relevance determination (ARD) is the standard device for input
selection in Gaussian process (GP) regression. By giving the covariance kernel a
separate lengthscale for every input and learning those lengthscales by
maximizing the marginal likelihood, ARD lets the data decide which coordinates
matter: irrelevant inputs receive very large lengthscales and are effectively
switched off. We trace this mechanism to the Bayesian Occam's razor embodied in
the marginal likelihood, derive the gradient through which it prunes inputs, and
emphasize that ARD delivers effective rather than exact sparsity. We review the
algorithms used in practice and the rules that turn lengthscales into selections,
and we survey the asymptotic theory, distinguishing the fixed-domain
identifiability obstruction on the lengthscales from the high-dimensional
selection-consistency guarantees recently established for hierarchical GP priors,
and noting what remains open for plain ARD. We compare ARD with spike-and-slab
priors, sparse axis-aligned and global-local shrinkage priors including the
Bayesian lasso and horseshoe, penalized-likelihood kriging, sensitivity and
projection criteria, and additive kernels. We argue that ARD endures because of
its seamless integration with kernel learning, universal software support, and
low cost, and we close with its limitations and remedies.
\end{abstract}


\bigskip

\section{Introduction and Historical Context}
\label{sec:intro}

Variable selection---deciding which of many candidate inputs actually influence
a response---is among the oldest and most consequential problems in statistics.
In linear models the question has a mature theory and a standard toolbox, from
subset selection to the lasso \citep{tibshirani1996} and its many descendants.
Outside the linear world the problem is harder: when the regression function may
bend, saturate, and interact in unknown ways, ``relevance'' is no longer a
matter of a single coefficient being nonzero, and the selection machinery must
be rebuilt on a nonparametric foundation. The difficulty is not merely technical.
In the linear model the meaning of ``input $j$ is irrelevant'' is unambiguous---
its coefficient is zero---but for a general function $f(x_1,\dots,x_D)$ one must
first decide what irrelevance \emph{means}: that $f$ is constant in $x_j$ for all
values of the other inputs, the natural and stringent definition we adopt
throughout, is a statement about a whole function rather than a single number, and
testing or enforcing it requires machinery that the linear toolbox does not
supply.

Gaussian process (GP) regression \citep{rw2006} is one of the most successful
nonparametric foundations available. A GP places a prior directly on the
unknown function, encodes smoothness and scale through a covariance kernel, and
delivers a full posterior---point predictions \emph{and} calibrated
uncertainty---in closed form for Gaussian noise. GPs are the workhorse of
computer-experiment emulation \citep{sacks1989,kennedy2001,santner2018}, are
ubiquitous in spatial statistics and machine learning, and increasingly drive
Bayesian optimization and the design of expensive experiments. In all of these
settings the analyst frequently confronts a list of inputs much longer than the
list that truly matters, and asks the GP not only to predict but to reveal which
inputs are doing the work.

To fix ideas and notation at the outset, we record the model that the whole
survey concerns. We observe pairs $\{(\bx_i, y_i)\}_{i=1}^N$ with inputs
$\bx_i \in \R^D$ and scalar responses, and posit the nonparametric regression
model
\begin{equation}
\label{eq:intromodel}
y_i = f(\bx_i) + \varepsilon_i,
\qquad \varepsilon_i \overset{\mathrm{iid}}{\sim} \Normal(0,\sigma_n^2),
\qquad f \sim \GP(0, k_{\btheta}),
\end{equation}
in which the unknown function $f$ is assigned a Gaussian process prior with
mean zero and covariance function $k_{\btheta}$ indexed by hyperparameters
$\btheta$. Section~\ref{sec:gpbackground} unpacks what \eqref{eq:intromodel}
buys; for now the point is only that all of the modeling freedom---and, as we
shall see, all of the variable selection---lives in the kernel $k_{\btheta}$ and
its hyperparameters. Although we develop the ideas for the Gaussian-noise
regression model \eqref{eq:intromodel}, which is where ARD is most transparent,
the same construction extends to categorical and count responses and to survival
outcomes within a generalized framework \citep{savitsky2011}, a breadth we
return to in Sections~\ref{sec:compare} and~\ref{sec:applications}.

\emph{Automatic relevance determination} (ARD) is the answer that the GP
literature reaches for first. The idea is disarmingly simple. The most common GP
kernels measure the similarity of two input points through a weighted distance;
ARD gives each input coordinate its own weight---its own
\emph{lengthscale}---and then \emph{learns} those lengthscales from data by
optimizing the GP's marginal likelihood. An input whose lengthscale is driven to
a very large value contributes almost nothing to the covariance, so the fitted
function is nearly flat along that coordinate and the input is, for practical
purposes, removed. The reciprocal of the lengthscale therefore behaves like a
relevance score, and reading off these scores constitutes a form of variable
selection that requires no separate selection algorithm, no tuning parameter set
by cross-validation, and essentially no extra code beyond switching a single
isotropic lengthscale for a vector of them.

The plan of the survey follows the logical order of the subject. After the
historical sketch that closes this section, Section~\ref{sec:why} develops the
\emph{theoretical foundations} of ARD: the kernel, the marginal likelihood, the
Bayesian Occam's razor that drives the mechanism, and the matrix-calculus
gradient through which that mechanism operates, with care taken to distinguish
the \emph{effective} sparsity ARD actually delivers from the \emph{exact}
sparsity that selection in the strict sense requires. Section~\ref{sec:algorithms}
then makes the method concrete, giving the algorithms---evidence maximization,
fully Bayesian sampling, variational approximation---and the thresholding rules
that convert lengthscales into selections. With the method and its computation in
hand, Section~\ref{sec:theory} establishes the \emph{asymptotic theory}: the
identifiability obstruction of fixed-domain asymptotics and the high-dimensional
selection-consistency results now available for hierarchical GP priors.
Section~\ref{sec:proscons} weighs ARD's advantages, disadvantages, and failure
modes in light of that theory, and Section~\ref{sec:compare} situates ARD within
the broader \emph{frontier landscape} of GP variable-selection methods---grouped
into probabilistic exact-sparsity priors, penalized-likelihood approaches, and
projection and sensitivity criteria---asking why, despite a crowded field of more
principled competitors, ARD remains the default. Section~\ref{sec:comput} treats
\emph{scalability} to large sample sizes and large input dimension, and
Section~\ref{sec:applications} surveys evaluation frameworks and application
domains. Section~\ref{sec:discussion} offers a discussion and outlook. Our
intended reader is a statistician who knows regression and a little about kernels
but who has not necessarily followed the GP literature; our emphasis throughout
is on intuition, with the heavier formulas and proof sketches set in displays
that can be skimmed without losing the argument.

A brief word on what this survey is \emph{not}. It is not a tutorial on Gaussian
processes in general---for that, \citet{rw2006} remains the definitive
reference---and it is not a comprehensive treatment of high-dimensional
nonparametric regression. It is a focused account of one idea, ARD, and of the
neighborhood of methods that surround it. We have tried to be candid about
ARD's failures as well as its successes, because the most useful thing a survey
can do for a practitioner is to say not only how a method works but when it
should not be trusted.

\subsection{A motivating illustration}
\label{sec:motivation}

Before any formalism, consider what ARD does on a problem where the truth is
known. We generated $N=120$ observations from a function of two inputs,
$y = g(x_1) + \varepsilon$, in which only the first input enters; the second
input $x_2$ is inert, varying over the same range but never affecting the
response. We fitted a GP with a squared-exponential ARD kernel
(Section~\ref{sec:why}) and optimized its two lengthscales by maximizing the
marginal likelihood. The fitted lengthscale for the relevant input came out near
$\hat\ell_1 \approx 1.7$, a moderate value reflecting genuine curvature; the
lengthscale for the inert input inflated to $\hat\ell_2 \approx 1.4\times 10^2$,
nearly two orders of magnitude larger.

\begin{figure}[t]
\centering
\includegraphics[width=0.75\textwidth]{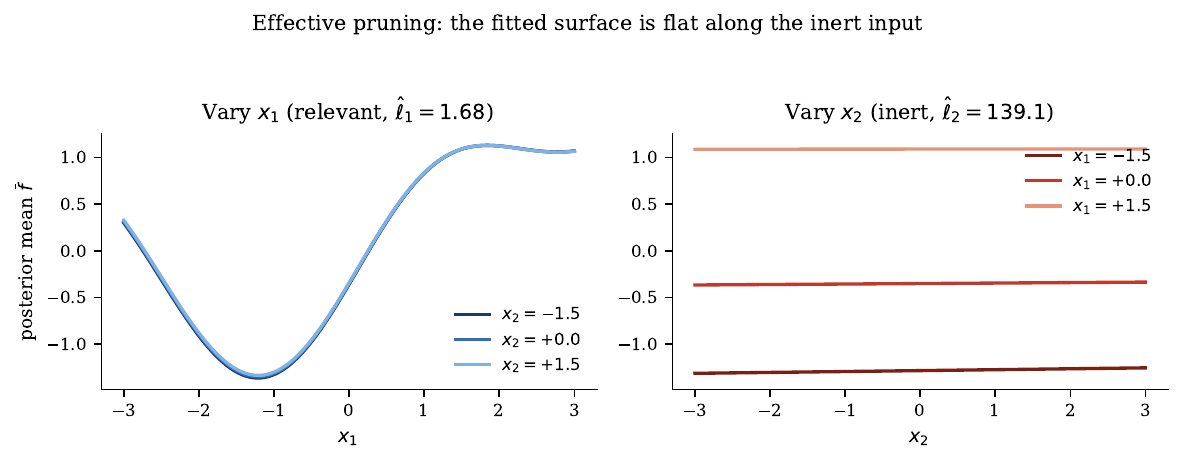}
\caption{Effective pruning by ARD on a two-input problem in which only $x_1$ is
relevant. After maximizing the marginal likelihood, the relevant input receives
a moderate lengthscale ($\hat\ell_1\approx 1.7$) and the posterior mean varies
with it (left), while the inert input receives a very large lengthscale
($\hat\ell_2\approx 1.4\times 10^2$) and the posterior mean is essentially flat
along it (right). The inert coordinate has been switched off without any
explicit selection step.}
\label{fig:lengthscale}
\end{figure}

Figure~\ref{fig:lengthscale} shows the consequence. Sweeping $x_1$ while holding
$x_2$ fixed traces out the underlying nonlinear signal (left panel); sweeping
$x_2$ while holding $x_1$ fixed produces nearly flat lines (right panel),
regardless of where $x_1$ is fixed. The model has \emph{discovered}, with no
guidance, that the response does not depend on $x_2$, and it has encoded that
discovery as a near-infinite lengthscale. This is ARD in a nutshell: relevance
becomes a learned property of the covariance, and selection becomes a by-product
of fitting.

The rest of this paper is, in a sense, an extended commentary on this figure:
why the inflation happens, when it can be trusted, how to read it as a selection
rule, and what to do when it fails.

\subsection{Historical development}
\label{sec:history}

ARD did not originate with Gaussian processes; it arrived in the GP literature
by a route that is worth tracing, both because the history clarifies what the
method is and because several of its modern variants are best understood as
returns to its origins.

The idea and the name are due to David MacKay, in the context of Bayesian neural
networks. In a Bayesian neural network each weight is given a Gaussian prior, and
MacKay's insight was to let the \emph{prior variance} of the weights emanating
from each input be a separate hyperparameter, controlled by its own
regularization constant, and to set those constants by maximizing the evidence
\citep{mackay1992occam}. An input irrelevant to the response would have its
weights pulled toward zero by a large regularization constant---its relevance
``determined automatically'' by the data. \citet{mackay1994} deployed exactly
this construction in the 1993 energy-prediction competition organized by ASHRAE,
a forecasting contest with roughly 150 entrants, and the ARD entry won by a
substantial margin. The episode established ARD as a practical tool and not
merely a theoretical nicety.

In parallel, Radford Neal was developing the Bayesian treatment of neural
networks that would become the standard reference \citep{neal1996}. Neal's
monograph did two things that matter for our story. First, it implemented ARD
priors in a fully Bayesian framework, sampling the relevance hyperparameters by
Markov chain Monte Carlo rather than fixing them at evidence-maximizing values---
the direct ancestor of the fully Bayesian ARD of Section~\ref{sec:fullbayes}.
Second, and more profoundly, Neal showed that a neural network with a single
hidden layer converges, as the number of hidden units tends to infinity and
under a suitable scaling of the priors, to a Gaussian process. This limit
theorem is the bridge: it revealed that the flexible Bayesian neural networks in
which ARD had been developed were, in a well-defined limit, Gaussian processes,
and it thereby licensed the transfer of ARD to GPs directly. Once one works with
the GP limit, the per-input regularization constants of the neural network become
the per-input lengthscales of the covariance kernel, and ARD takes the form
\eqref{eq:seard} we use today.

The GP form was brought into mainstream machine learning by
\citet{williams1996}, who fitted GP regression with per-dimension lengthscales,
and it received its canonical exposition in the monograph of \citet{rw2006},
whose Chapter~5 on model selection presents the ARD kernel and the
evidence-maximization framework in the notation that has become standard and
that we follow here.

A related but distinct lineage deserves mention to forestall a common confusion.
Michael Tipping's relevance vector machine \citep{tipping2001} applies an
ARD-style prior not to input dimensions but to \emph{basis functions} or
\emph{training points}, giving each its own prior precision and pruning those
whose precision diverges. The result is sparsity in the \emph{representation}---a
model supported on a few ``relevance vectors''---rather than selection among
input variables. The mathematics rhymes with input ARD, and the two are often
discussed together under the heading of sparse Bayesian learning, but they answer
different questions: the relevance vector machine asks which \emph{data points} to
keep, while input ARD asks which \emph{inputs} matter. \citet{wipf2008} later gave
a unifying theoretical analysis of this whole family, showing among other things
that the ARD objective can be recast as a series of reweighted $\ell_1$ problems
and is equivalent to a particular non-factorial, data-dependent prior---an
analysis that also exposed the non-convexity we discuss in
Section~\ref{sec:landscape}. \citet{qi2004} contributed a further refinement,
observing that evidence-maximizing ARD can overfit and proposing a predictive
variant that selects relevances by leave-one-out predictive performance. The
sparse-Bayesian-learning program remains active: \citet{helgoy2025}, for
instance, develop a Bayesian-lasso-based sparse learning model trained by the same
evidence-maximization principle and benchmark it directly against the relevance
vector machine, a reminder that the type-II machinery underlying ARD is still
being refined.

With this lineage in place---MacKay's evidence-based regularization, Neal's
Bayesian sampling and GP limit, the Rasmussen--Williams synthesis, and the sparse
Bayesian learning analysis---the modern landscape of Section~\ref{sec:compare}
falls into order: the fully Bayesian methods return to Neal, the penalized
methods make Wipf and Nagarajan's $\ell_1$ connection explicit, and the
predictive and sensitivity methods answer Qi and colleagues' worry that the
evidence is not quite the right objective.

\section{Theoretical Foundations of ARD}
\label{sec:why}

\subsection{Gaussian process regression in one page}
\label{sec:gpbackground}

We observe data $\{(\bx_i, y_i)\}_{i=1}^N$ with inputs $\bx_i \in \R^D$ and
scalar responses $y_i$, and posit
\begin{equation}
\label{eq:model}
y_i = f(\bx_i) + \varepsilon_i, \qquad
\varepsilon_i \overset{\mathrm{iid}}{\sim} \Normal(0, \sigma_n^2),
\end{equation}
where the unknown function $f$ is given a Gaussian process prior,
$f \sim \GP(0, k_{\btheta})$, with mean zero (after centering) and covariance
function $k_{\btheta}(\bx, \bx')$ indexed by hyperparameters $\btheta$. The
defining property of a GP is that the function values at any finite set of
inputs are jointly Gaussian. Collecting the training inputs into $X$ and writing
$\Kff \in \R^{N\times N}$ for the matrix with entries
$[\Kff]_{ij} = k_{\btheta}(\bx_i, \bx_j)$, the prior over the training function
values is $\bff \sim \Normal(\mathbf{0}, \Kff)$, and the responses satisfy
$\by \sim \Normal(\mathbf{0}, \Ky)$ with $\Ky = \Kff + \sigma_n^2 I$.

For a test input $\bx_\ast$, joint Gaussianity gives a closed-form posterior
predictive distribution,
\begin{equation}
\label{eq:posterior}
f(\bx_\ast)\mid \by
\sim \Normal\!\big(\,\bar f(\bx_\ast),\ \mathbb{V}[f(\bx_\ast)]\,\big),
\end{equation}
with mean and variance
\begin{align}
\bar f(\bx_\ast)
  &= \mathbf{k}_\ast^\top \Ky^{-1}\by, \label{eq:postmean}\\
\mathbb{V}[f(\bx_\ast)]
  &= k_{\btheta}(\bx_\ast, \bx_\ast)
   - \mathbf{k}_\ast^\top \Ky^{-1}\mathbf{k}_\ast, \label{eq:postvar}
\end{align}
where $\mathbf{k}_\ast = (k_{\btheta}(\bx_\ast,\bx_1), \dots,
k_{\btheta}(\bx_\ast,\bx_N))^\top$. Equations~\eqref{eq:postmean}
and~\eqref{eq:postvar} are the entire predictive engine of GP regression: the
mean is a smooth interpolant of the data and the variance grows away from the
observations, providing the uncertainty quantification that makes GPs attractive.

Everything now hinges on the kernel and its hyperparameters. The kernel encodes
what we believe about $f$---how smooth it is, how quickly it varies, and, for
our purposes, \emph{which inputs it varies with}.

\subsection{The ARD kernel}
\label{sec:ardkernel}

The most common stationary kernel is the squared-exponential (SE), also called
the radial-basis-function or exponentiated-quadratic kernel. In its
\emph{isotropic} form it uses a single lengthscale $\ell$ for all inputs:
$k(\bx,\bx') = \sigma_f^2 \exp(-\|\bx-\bx'\|^2 / 2\ell^2)$. This is rigid: it
insists that the function vary at the same rate in every direction. The ARD
generalization \citep{mackay1994,neal1996,rw2006} replaces the single
lengthscale with one lengthscale per input dimension,
\begin{equation}
\label{eq:seard}
\begin{split}
k_{\btheta}(\bx, \bx')
&= \sigma_f^2 \exp\!\left(
   -\frac{1}{2}\sum_{j=1}^D \frac{(x_j - x'_j)^2}{\ell_j^2}
  \right)\\
&= \sigma_f^2 \exp\!\left(
   -\tfrac{1}{2}(\bx-\bx')^\top M (\bx-\bx')
  \right),
\end{split}
\end{equation}
with $M = \diag(\ell_1^{-2}, \dots, \ell_D^{-2})$ and hyperparameters
$\btheta = (\ell_1, \dots, \ell_D, \sigma_f, \sigma_n)$. The signal variance
$\sigma_f^2$ sets the overall amplitude and $\sigma_n^2$ the noise level. The
name ``automatic relevance determination'' is due to \citet{mackay1994}, who
introduced the device in the context of Bayesian neural networks; the same
structure was developed in parallel by \citet{neal1996}, whose demonstration
that a one-hidden-layer Bayesian neural network converges to a Gaussian process
as the number of hidden units grows supplied the bridge by which ARD passed from
neural networks into the GP literature.

The crucial feature of \eqref{eq:seard} is that the kernel \emph{factorizes}
across input dimensions: the exponent is a sum of per-coordinate contributions,
so the kernel is a product
$k_{\btheta}(\bx,\bx') = \sigma_f^2 \prod_{j=1}^D
\exp\!\big(-(x_j-x'_j)^2/2\ell_j^2\big)$ of one-dimensional factors. Consider the
$j$th factor as $\ell_j \to \infty$. For any fixed pair of inputs the difference
$x_j - x'_j$ is finite, so $(x_j-x'_j)^2/\ell_j^2 \to 0$ and the factor tends to
$1$. A factor identically equal to $1$ contributes nothing to the product: the
covariance between any two points becomes independent of their $j$th
coordinates. Because the covariance controls the function through
\eqref{eq:postmean}, the posterior mean becomes constant in the $j$th input. In
the language of selection, the input has been \emph{pruned}.

It is therefore natural to read the reciprocal lengthscale $1/\ell_j$ as the
\emph{relevance} of input $j$: a large $1/\ell_j$ means the function changes
rapidly along coordinate $j$ (high relevance), while $1/\ell_j \approx 0$ means
the function is nearly flat along it (irrelevance). The matrix $M$ in
\eqref{eq:seard} is the diagonal special case of a general positive-definite
$M$; allowing $M$ to be a low-rank-plus-diagonal or full matrix yields
\emph{factor analysis} and \emph{linear-projection} variants
\citep[][\S5.1]{rw2006} that select linear combinations of inputs rather than
individual coordinates, a theme we return to in
Section~\ref{sec:compare} under active subspaces.

ARD is not tied to the SE kernel. The Mat\'ern family, often preferred in
spatial statistics and increasingly in machine learning because it produces
rougher, more realistic sample paths \citep{stein1999}, admits the same
per-dimension treatment. The Mat\'ern-$\nu$ ARD kernel replaces the Euclidean
distance by the ARD-weighted distance
$r(\bx,\bx') = \big(\sum_j (x_j-x'_j)^2/\ell_j^2\big)^{1/2}$ inside the usual
Mat\'ern form; for the common choices $\nu = 3/2$ and $\nu = 5/2$ this gives,
respectively, $k = \sigma_f^2(1+\sqrt{3}\,r)\exp(-\sqrt{3}\,r)$ and
$k = \sigma_f^2(1 + \sqrt{5}\,r + \tfrac{5}{3}r^2)\exp(-\sqrt{5}\,r)$. The
relevance interpretation of $1/\ell_j$ carries over unchanged.

\subsection{The marginal likelihood and the Bayesian Occam's razor}
\label{sec:evidence}

If lengthscales encode relevance, how are they chosen? The answer that defines
ARD is \emph{Type-II maximum likelihood}, also called \emph{evidence
maximization}: one selects $\btheta$ by maximizing the marginal likelihood of
the data, obtained by integrating the latent function out of the model. Because
the prior and the noise are Gaussian, the integral is available in closed form,
and its logarithm is
\begin{equation}
\label{eq:logml}
\log p(\by\mid X, \btheta)
= \underbrace{-\tfrac{1}{2}\,\by^\top \Ky^{-1}\by}_{\text{data fit}}
\;\underbrace{-\,\tfrac{1}{2}\log|\Ky|}_{\text{complexity penalty}}
\;-\;\frac{N}{2}\log 2\pi .
\end{equation}
The decomposition annotated in \eqref{eq:logml} is the key to understanding ARD,
and it deserves to be read slowly. The first term, $-\tfrac12 \by^\top
\Ky^{-1}\by$, is a data-fit term: it is large (close to zero from below) when the
covariance model explains the observed responses well. The second term,
$-\tfrac12\log|\Ky|$, is a complexity penalty: the log-determinant of the
covariance grows as the model is made more flexible, and the minus sign means
that flexibility is penalized. The third term is a constant.

This is the \emph{Bayesian Occam's razor} \citep{mackay1992occam,
rasmussen2001occam} made arithmetic. The marginal likelihood is a normalized
probability distribution over data sets: $\int p(\by\mid X,\btheta)\,d\by = 1$. A
model flexible enough to fit many possible data sets must, by conservation of
probability, assign a lower density to any particular one; a model too rigid to
fit the data at hand assigns it low density for a different reason. The
marginal likelihood is maximized in between, by a model just flexible enough to
capture the signal and no more. Crucially, this trade-off is \emph{automatic}:
it falls out of integrating over the latent function, with no penalty term
inserted by hand and no tuning constant to set.

It is worth dwelling on why this differs from ordinary maximum likelihood, since
the distinction is the whole point. If we had \emph{not} integrated out the
latent function---if we had instead maximized the likelihood of the data over
both $f$ and $\btheta$ jointly---there would be no complexity penalty at all, and
the optimizer would drive every lengthscale to zero, fitting the noise. The
penalty in \eqref{eq:logml} arises precisely because $f$ has been marginalized:
the term $-\tfrac12\log|\Ky|$ is the log-volume of the space of functions the
prior can represent at the given hyperparameters, and a larger such volume is
penalized because the prior probability must be spread more thinly. Type-II
maximum likelihood is thus maximum likelihood \emph{after} the right integration,
and it is that integration that supplies the razor. The same logic underlies the
use of the marginal likelihood for choosing among entirely different models, not
just among hyperparameters \citep{rasmussen2001occam}, and it is one of the most
attractive features of the Bayesian framework that the penalty is not a separate
ingredient but a consequence of the probability calculus.

\begin{figure}[t]
\centering
\includegraphics[width=0.60\textwidth]{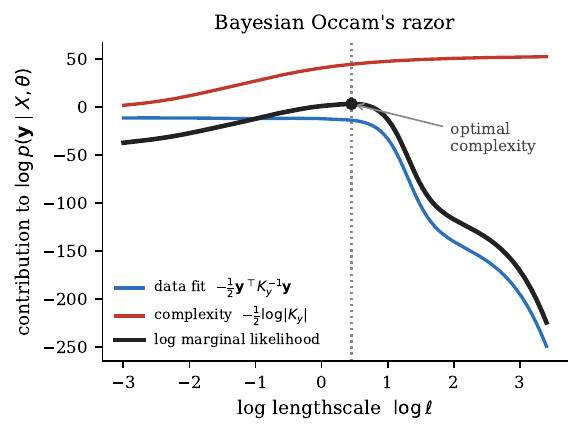}
\caption{The two competing terms of the log marginal likelihood
\eqref{eq:logml}, computed for a real one-dimensional GP, as functions of the
log lengthscale. Shrinking the lengthscale improves the data fit (blue) but
incurs a larger complexity penalty (red); the marginal likelihood (black) peaks
at an intermediate lengthscale that balances the two. This automatic balancing
is the Bayesian Occam's razor, and it is the engine that drives irrelevant
inputs toward large lengthscales.}
\label{fig:evidence}
\end{figure}

Now apply the razor to a single lengthscale. Figure~\ref{fig:evidence} traces
the two terms of \eqref{eq:logml} as a function of $\log\ell$ for a real
one-dimensional GP fit. As $\ell$ decreases the model becomes more flexible: the
data-fit term improves (rises) because a wiggly function can thread the data
more closely, but the complexity penalty worsens (falls) because a flexible
covariance can fit many data sets. As $\ell$ increases the reverse happens. The
marginal likelihood---the sum---peaks at an interior, intermediate lengthscale,
the one that balances fit against complexity.

The mechanism that prunes an \emph{irrelevant} input now follows directly.
Suppose coordinate $j$ genuinely does not affect the response. Giving it a
finite lengthscale only adds flexibility---the freedom to fit spurious variation
along a direction in which the truth is flat---and that flexibility is penalized
by the complexity term without any compensating gain in the data-fit term, since
there is no real signal to capture. The marginal likelihood is therefore
increased by sending $\ell_j$ toward infinity, removing the unhelpful
flexibility. Conversely, for a relevant input the data-fit gain from a finite
lengthscale outweighs the complexity cost, and the optimizer settles on a
moderate value. This is exactly the asymmetry we saw numerically in
Figure~\ref{fig:lengthscale}: the relevant input kept a moderate lengthscale,
the inert input inflated.

\subsection{The gradient of the marginal likelihood}
\label{sec:gradient}

The intuition above becomes precise once we write down how the marginal
likelihood changes as a lengthscale moves, and the derivation repays the small
amount of matrix calculus it requires, because it exhibits the pruning mechanism
as an explicit competition between two trace terms. We state the gradient and
sketch its derivation; the result is standard \citep[][Eq.~5.9]{rw2006}, but
seeing the mechanics clarifies exactly how the complexity penalty acts during
optimization.

\begin{proofsketch}
Write $\mathcal{L}(\btheta) = \log p(\by\mid X,\btheta)$ as in \eqref{eq:logml}.
Two matrix differentials do the work. For a parameter-dependent invertible matrix
$\Ky$, differentiating the identity $\Ky \Ky^{-1} = I$ gives the differential of
the inverse,
\begin{equation}
\label{eq:dinv}
d(\Ky^{-1}) = -\,\Ky^{-1}\,(d\Ky)\,\Ky^{-1},
\end{equation}
while Jacobi's formula gives the differential of the log-determinant,
\begin{equation}
\label{eq:dlogdet}
d\,\log|\Ky| = \tr\!\big(\Ky^{-1}\,d\Ky\big).
\end{equation}
Applying \eqref{eq:dinv} and \eqref{eq:dlogdet} to the data-fit and
complexity terms of \eqref{eq:logml},
\[
d\mathcal{L}
= \tfrac{1}{2}\,\by^\top \Ky^{-1}(d\Ky)\Ky^{-1}\by
 - \tfrac{1}{2}\tr\!\big(\Ky^{-1}\,d\Ky\big).
\]
The first term is a scalar, hence equal to its own trace; writing
$\balpha = \Ky^{-1}\by$ and using the cyclic property of the trace turns it into
$\tfrac{1}{2}\tr(\balpha\balpha^\top d\Ky)$. Combining the two terms,
$d\mathcal{L} = \tfrac{1}{2}\tr[(\balpha\balpha^\top - \Ky^{-1})\,d\Ky]$, and
reading off the coefficient of $d\theta_m$ yields the gradient.
\end{proofsketch}

\noindent The result is
\begin{equation}
\label{eq:grad}
\begin{split}
\frac{\partial}{\partial \theta_m}\log p(\by\mid X,\btheta)
&= \tfrac{1}{2}\,\tr\!\left[
   \big(\balpha\balpha^\top - \Ky^{-1}\big)
   \frac{\partial \Ky}{\partial \theta_m}
  \right],\\
&\qquad\qquad \balpha = \Ky^{-1}\by .
\end{split}
\end{equation}
The two pieces are exactly the two sides of the Occam's razor. The term
$\balpha\balpha^\top$ is the data-fit gradient: it rewards changes in $\Ky$ that
better explain the residuals. The term $\Ky^{-1}$ is the complexity gradient: it
penalizes changes that make the model more flexible. A hyperparameter sits at a
stationary point when these balance, $\tr(\balpha\balpha^\top \partial_m \Ky)
= \tr(\Ky^{-1}\partial_m \Ky)$, which is the differential form of the fit-versus-
complexity trade-off of Section~\ref{sec:evidence}.

To see the pruning concretely, specialize to the SE-ARD kernel \eqref{eq:seard}.
The derivative of the covariance with respect to the $d$th lengthscale has entries
\begin{equation}
\label{eq:dkdl}
\left[\frac{\partial \Kff}{\partial \ell_d}\right]_{ij}
= [\Kff]_{ij}\,\frac{(x_{di}-x_{dj})^2}{\ell_d^{3}} ,
\end{equation}
with analogous closed forms for $\sigma_f$ and $\sigma_n$. Now suppose input $d$
is irrelevant. Then the residual structure carries no systematic variation along
coordinate $d$, so the data-fit term $\tfrac12 \tr(\balpha\balpha^\top \partial_{\ell_d}\Kff)$
is strongly attenuated; meanwhile the complexity term
$ - \tfrac12\tr(\Ky^{-1}\partial_{\ell_d}\Kff)$ is positive for the squared-exponential kernel. The gradient
\eqref{eq:grad} is therefore negative in the direction of decreasing $\ell_d$:
the optimizer increases $\ell_d$, and \eqref{eq:dkdl} shows the force driving it
weakens  as $\ell_d$ grows (the entries decay like $\ell_d^{-3}$), so the lengthscale climbs to a large but finite plateau where the input is effectively pruned. This is the
quantitative content of the heuristic in Section~\ref{sec:evidence}, and it is
the engine that produced the order-of-magnitude lengthscale separation in
Figure~\ref{fig:lengthscale}. The same gradient \eqref{eq:grad} is what every
optimizer in Section~\ref{sec:algorithms} ascends.

\subsection{Effective sparsity, not exact sparsity}
\label{sec:effectivesparsity}

It is essential to be precise about what kind of sparsity ARD delivers, because
loose language here is the source of much confusion. Evidence maximization drives
the lengthscales of irrelevant inputs to \emph{large} values, not to infinity.
In the illustration of Section~\ref{sec:motivation} the inert lengthscale
reached order $10^2$, which makes the input negligible but not formally absent;
in a finite sample it will essentially never be exactly $+\infty$, and the
corresponding relevance $1/\ell_j$ will be small but positive. ARD thus produces
\emph{effective} or \emph{soft} sparsity: a ranking of inputs by relevance in
which the irrelevant ones cluster near zero, rather than \emph{exact} sparsity
in which a subset of relevances are identically zero.

\begin{figure}[t]
\centering
\includegraphics[width=0.62\textwidth]{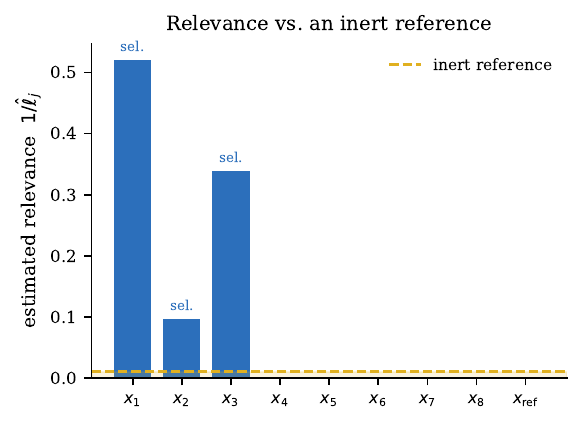}
\caption{Estimated relevances $1/\hat\ell_j$ for an eight-input problem in which
$x_1, x_2, x_3$ are relevant (with genuinely different strengths) and
$x_4,\dots,x_8$ are inert, fitted by evidence maximization. An artificial inert
``reference'' input $x_{\mathrm{ref}}$ calibrates the noise floor (shaded band):
inputs whose relevance clearly exceeds it are selected. ARD ranks the inputs
correctly here, but the selection boundary itself must be supplied by a
thresholding rule.}
\label{fig:relevance}
\end{figure}

This distinction has two practical consequences that run through the rest of the
survey. First, ARD on its own does not return a selected subset; it returns a
relevance vector. Converting that vector into a yes/no decision for each input
requires a separate \emph{thresholding rule}, and the choice of rule is
consequential and somewhat arbitrary (Section~\ref{sec:thresholding}).
Figure~\ref{fig:relevance} illustrates this on an eight-input problem in which
three inputs are relevant and five are inert: the relevant inputs stand clearly
above the inert ones, but the boundary between ``selected'' and ``not selected''
is a line the analyst must draw. A useful device, due to \citet{linkletter2006}
and discussed in Section~\ref{sec:compare}, is to draw that line using an
artificially constructed input known to be inert, whose estimated relevance
calibrates the noise floor.

Second, the soft nature of ARD sparsity is the precise point at which more
elaborate methods enter. Spike-and-slab priors (Section~\ref{sec:spikeslab})
build exact zeros into the model through a discrete inclusion indicator;
penalized-likelihood approaches (Section~\ref{sec:penalized}) borrow the lasso's
ability to set parameters exactly to zero. Both can be seen as responses to the
fact that ARD, left to itself, only ranks.

\subsection{A cautionary subtlety: the landscape is not convex}
\label{sec:landscape}

A final point completes the picture of \emph{why} ARD works---and warns of when
it does not. The objective \eqref{eq:logml} is, as a function of the
hyperparameters, generally \emph{non-convex} and may be \emph{multimodal}.
\citet{wipf2008}, analyzing the closely related sparse Bayesian learning
problem, made the structure explicit: the data-fit term is convex in the
relevance parameters while the log-determinant term is concave, so their sum can
have multiple local optima and saddle points. In a GP with $D$ lengthscales the
optimizer navigates a $(D+2)$-dimensional surface that may have several basins,
each corresponding to a different pattern of which inputs are deemed relevant.

The consequence is that the relevance pattern ARD reports can depend on where
the optimization starts. Two practitioners running the same model from different
initializations may converge to different lengthscale vectors and hence different
selections. This is not a flaw in the Occam's-razor logic of
Section~\ref{sec:evidence}---that logic correctly identifies the \emph{global}
optimum as the right balance of fit and complexity---but a flaw in our ability to
\emph{find} it. It motivates the multi-start strategies and the fully Bayesian
alternatives that we take up in Sections~\ref{sec:algorithms}
and~\ref{sec:compare}, and it is one of the recurring reasons that ARD's
relevances should be read as informative but not infallible.

\section{Algorithmic Implementations and Thresholding}
\label{sec:algorithms}

This section assembles the concrete procedures by which ARD models are fitted and
turned into variable-selection decisions. We organize them as three inference
strategies---evidence maximization, fully Bayesian sampling, and variational
approximation---followed by the thresholding rules that all three require, and a
note on software. The aim is to give the reader enough algorithmic detail to know
what each library is doing under the hood and to implement the methods if needed.

\subsection{Type-II maximum likelihood by gradient ascent}
\label{sec:typeII}

The canonical ARD algorithm maximizes the log marginal likelihood
\eqref{eq:logml} over the hyperparameter vector $\btheta$ by
gradient-based optimization, using the analytic gradient
\eqref{eq:grad}--\eqref{eq:dkdl}. Two practical refinements are universal. First,
the positive hyperparameters are optimized on the \emph{log scale}, both to
enforce positivity and to make the landscape better behaved; the chain rule turns
\eqref{eq:grad} into a gradient with respect to $\log\theta_m$ by a factor of
$\theta_m$. Second, because the surface is non-convex
(Section~\ref{sec:landscape}), the optimization is run from \emph{multiple random
restarts} and the solution with the highest marginal likelihood is kept.
Algorithm~\ref{alg:typeII} states the procedure.

\begin{algorithm}[t]
\caption{ARD by Type-II maximum likelihood (evidence maximization)}
\label{alg:typeII}
\begin{algorithmic}[1]
\State \textbf{Input:} data $(X, \by)$ with $\by$ centered; number of restarts $R$
\State \textbf{Output:} fitted hyperparameters $\hat\btheta$, relevances $1/\hat\ell_j$
\State Standardize each input column of $X$ to zero mean and unit variance
\For{$r = 1, \dots, R$}
  \State Initialize $\log\ell_j$ (e.g., near $\log(\text{median pairwise distance})$),
         $\log\sigma_f$, $\log\sigma_n$
  \Repeat
    \State Form $\Ky = \Kff + \sigma_n^2 I$ using the ARD kernel \eqref{eq:seard}
    \State Compute Cholesky $\Ky = LL^\top$ \Comment{$O(N^3)$}
    \State Solve $\balpha = L^{-\top}(L^{-1}\by)$
    \State Evaluate $\log p(\by\mid X,\btheta)$ via \eqref{eq:logml}
    \State For each $\theta_m$, form $\partial\Ky/\partial\theta_m$ and the
           gradient \eqref{eq:grad} \Comment{$O(N^2 D)$}
    \State Take an L-BFGS / conjugate-gradient step on $\log\btheta$
  \Until{convergence}
  \State Record $\btheta^{(r)}$ and its marginal likelihood
\EndFor
\State $\hat\btheta \gets \argmax_r \log p(\by\mid X,\btheta^{(r)})$
\State \Return $\hat\btheta$ and relevances $1/\hat\ell_j$
\end{algorithmic}
\end{algorithm}

The optimizer of choice is a quasi-Newton method---L-BFGS is the most common---or
nonlinear conjugate gradients; \citet{rw2006} use conjugate gradients and a
related scaled-conjugate-gradient method is standard in the GPML toolbox. The cost
is dominated by the $O(N^3)$ factorization on line~8, repeated across iterations
and restarts as discussed in Section~\ref{sec:cubic}. For large $N$ the exact
factorization is replaced by one of the scalable approximations of
Section~\ref{sec:scalable}, with the rest of the algorithm essentially unchanged.

Several practical matters determine whether Algorithm~\ref{alg:typeII} succeeds in
practice, and they are worth stating because they are rarely written down. The
\emph{initialization} of the lengthscales matters a great deal given the
non-convexity: a common and effective heuristic sets each $\log\ell_j$ near the
log of the median pairwise distance along coordinate $j$, which places the
optimizer in a sensible region where the kernel is neither saturated nor
degenerate. \emph{Standardizing} the inputs to common scale, as on line~3, is not
cosmetic---without it the lengthscales must absorb the differing scales of the
inputs, which both slows the optimization and muddies the interpretation of
$1/\hat\ell_j$ as relevance, since a large lengthscale could then reflect either
irrelevance or a large input scale. The \emph{number of restarts} trades
computation against the risk of a poor local optimum; a handful suffices in low
dimension, but the requirement grows with $D$, and in high dimension the
diminishing returns of restarts are part of what motivates the fully Bayesian
alternatives that explore the surface rather than seeking a single mode.
\emph{Convergence} is typically judged by the change in the marginal likelihood
and the norm of its gradient, and it is good practice to inspect the spread of
the marginal likelihood across restarts: a wide spread is a warning that the
surface is badly multimodal and that the reported relevances should be treated
with corresponding caution. Finally, because the lengthscales are optimized on the
log scale, the optimizer naturally respects their positivity and explores the
many orders of magnitude that separate a relevant lengthscale from an inflated
one, as in the factor of nearly a hundred we saw in Section~\ref{sec:motivation}.

Because the surface is multimodal, multi-start gradient ascent is not the only
option, and it is worth noting that the marginal likelihood can be treated as a
black-box objective and handed to a general-purpose global optimizer for
hyperparameters---Bayesian optimization \citep{snoek2012} or tree-structured
Parzen estimators \citep{bergstra2011}, the standard tools of automated
hyperparameter search. These methods can sometimes escape poor basins that trap
local ascent, at the cost of more objective evaluations (each still paying the
factorization cost of Section~\ref{sec:cubic}). In practice they are used more
often to tune the surrounding modeling choices than to replace gradient ascent on
the lengthscales themselves, but they are a legitimate response to the
multimodality, and the fully Bayesian sampling of Section~\ref{sec:algo_bayes} can
be seen as the principled limit of the same impulse to explore the surface rather
than descend a single basin.

\subsection{Fully Bayesian inference by sampling}
\label{sec:algo_bayes}

The fully Bayesian treatment of Section~\ref{sec:fullbayes} replaces the
maximization with integration. One places a prior $p(\btheta)$ on the
hyperparameters---log-normal priors on the lengthscales are common, and the
half-Cauchy SAAS hierarchy of \citet{eriksson2021} when strong sparsity is
desired---and draws samples $\btheta^{(1)},\dots,\btheta^{(S)}$ from the posterior
$p(\btheta\mid \by) \propto p(\by\mid X,\btheta)\,p(\btheta)$.
Algorithm~\ref{alg:bayes} sketches the procedure with Hamiltonian Monte Carlo;
slice sampling \citep{murray2010} is a gradient-free alternative.

\begin{algorithm}[t]
\caption{Fully Bayesian ARD by Hamiltonian Monte Carlo}
\label{alg:bayes}
\begin{algorithmic}[1]
\State \textbf{Input:} data $(X,\by)$; prior $p(\btheta)$; number of samples $S$
\State \textbf{Output:} posterior samples of relevances; relevance summaries
\State Initialize $\btheta^{(0)}$
\For{$s = 1, \dots, S$}
  \State Propose $\btheta'$ by simulating Hamiltonian dynamics on
         $\log p(\by\mid X,\btheta) + \log p(\btheta)$,
         using gradient \eqref{eq:grad}
  \State Accept or reject $\btheta'$ by the Metropolis criterion
  \State Record $\btheta^{(s)}$ and the relevances $1/\ell_j^{(s)}$
\EndFor
\State Summarize each input by, e.g., the posterior mean of $1/\ell_j$ or the
       posterior probability that $1/\ell_j$ exceeds an inert-reference level
\State \Return relevance posteriors and summaries
\end{algorithmic}
\end{algorithm}

Relevance is then assessed from the posterior rather than from a point estimate:
one can report the posterior distribution of each $1/\ell_j$, the posterior
probability that it exceeds a reference level (Section~\ref{sec:thresholding}),
or, in a spike-and-slab formulation, the posterior inclusion probability directly.
The cost is $S$ times the per-iteration GP cost, with $S$ in the hundreds or
thousands, which is why this route is reserved for problems where the extra
fidelity---honest uncertainty, robust high-dimensional selection---justifies the
expense.

\subsection{Variational and approximate variants}
\label{sec:algo_var}

Between the point estimate and full MCMC sits a spectrum of approximate-inference
methods. The sparse variational GP of \citet{titsias2009} and \citet{hensman2013}
optimizes a lower bound on the marginal likelihood and treats the ARD lengthscales
as parameters of that bound, giving scalable Type-II-like estimation. The scalable
spike-and-slab variational GP of \citet{dance2022} adds variational distributions
over inclusion indicators, returning posterior inclusion probabilities at
SVGP-like cost---arguably the most practical route to calibrated GP variable
selection at scale. Expectation propagation underlies the predictive-ARD method of
\citet{qi2004}, which approximates the leave-one-out predictive distribution and
selects lengthscales to optimize predictive rather than marginal likelihood,
guarding against the evidence overfitting of Section~\ref{sec:dimburden}.

\subsection{From relevances to selections: thresholding rules}
\label{sec:thresholding}

Because ARD yields effective rather than exact sparsity
(Section~\ref{sec:effectivesparsity}), every variant above must convert a vector
of relevances into a discrete selection. The common rules are the following.
\begin{enumerate}
\item \emph{Fixed cutoff.} Declare input $j$ relevant if $1/\hat\ell_j$ exceeds a
fixed threshold, or equivalently if $\hat\ell_j$ is below a lengthscale ceiling.
Simple but sensitive to the scale of the problem and to the arbitrary cutoff.
\item \emph{Gap or elbow.} Sort the relevances and cut at the largest gap, on the
heuristic that relevant and irrelevant inputs form two clusters separated by a
visible jump. Effective when the separation is clean, as in
Figure~\ref{fig:relevance}, but ill-defined when relevances vary smoothly.
\item \emph{Inert-reference calibration.} Augment the inputs with one or more
artificial variables known to be irrelevant, fit the model, and use the
estimated relevance of the artificial variables as a noise floor; select inputs
whose relevance clearly exceeds it \citep{linkletter2006}. This is the device of
Figure~\ref{fig:relevance} and is more defensible than a fixed cutoff because it
calibrates to the data and model at hand. A permutation version, in which an input
is randomly permuted to destroy its relationship with the response, gives a related
reference distribution.
\item \emph{Posterior inclusion probability.} In a Bayesian spike-and-slab model,
select input $j$ if its posterior inclusion probability exceeds $1/2$ (the median
probability model) or another decision-theoretic threshold. This is the most
principled rule but requires the spike-and-slab machinery of
Section~\ref{sec:spikeslab} rather than plain ARD.
\item \emph{Knockoff filtering with error control.} The inert-reference and
permutation devices above are informal precursors of a rigorous idea: the
model-X knockoffs framework of \citet{candes2018} constructs synthetic
``knockoff'' variables that mimic the dependence structure of the real inputs but
are conditionally independent of the response, and uses them to select inputs with
\emph{finite-sample control of the false discovery rate}. Adapting this machinery
to kernel and nonparametric models is an active area; \citet{dai2023} integrate
the knockoff filter with random-feature approximations and subsampling to
guarantee false-discovery-rate control for nonparametric additive models, the
closest kernel-based analogue. Knockoff filtering is the natural choice when a
calibrated error guarantee on the selected set, rather than a mere ranking, is
required---a guarantee that raw lengthscale thresholding cannot offer.
\end{enumerate}
The dependence of the final selection on this choice is, we stress once more, a
genuine limitation of relevance-based selection, and it is the reason that methods
with built-in exact sparsity, inclusion probabilities, or error control are
preferable when the selection itself---rather than the resulting prediction---is
the object of inference.

\subsection{Software}
\label{sec:software}

All of the above is available off the shelf. GPy \citep{gpy2014}, GPflow
\citep{gpflow2017}, GPyTorch \citep{gardner2018}, the GPML toolbox of
\citet{rw2006}, and scikit-learn \citep{pedregosa2011} all provide ARD kernels and
marginal-likelihood optimization; GPflow and GPyTorch integrate with automatic
differentiation and GPU acceleration and support SVGP for scalability; fully
Bayesian inference is available through probabilistic-programming interfaces and
through the SAAS implementations released with \citet{eriksson2021}. For the
practitioner, fitting an ARD model and reading off relevances is genuinely a
matter of a few lines of code---which returns us, by way of the algorithms, to the
reason for ARD's popularity.

\subsection{A practical decision guide}
\label{sec:guide}

It may help to collect the foregoing into concrete guidance, organized by the two
quantities that most determine the right approach: the sample size $N$ and the
input dimension $D$.

\emph{Small $N$ (up to a few thousand), low to moderate $D$ (up to roughly
twenty).} This is ARD's home ground. Use an exact GP with an SE-ARD or
Mat\'ern-ARD kernel, fit by Type-II maximum likelihood
(Algorithm~\ref{alg:typeII}) with input standardization, median-distance
initialization, and a handful of restarts. Read off the relevance ranking, and
calibrate a selection threshold with one or more inert reference inputs
(Section~\ref{sec:thresholding}). Check the spread of the marginal likelihood
across restarts as a multimodality diagnostic. This recipe is fast, interpretable,
and usually correct.

\emph{Small $N$, high $D$ (many tens or more).} Here plain lengthscale
thresholding becomes unreliable (Section~\ref{sec:disadvantages},
Figure~\ref{fig:highdim}), and the extra cost of a more principled method is
warranted. Prefer a fully Bayesian treatment with a strong sparsity prior---the
SAAS half-Cauchy hierarchy \citep{eriksson2021} sampled by the No-U-Turn sampler
(Algorithm~\ref{alg:bayes})---or a spike-and-slab model for calibrated inclusion
probabilities. If interactions are of interest or the dimension is very high,
consider a low-order additive GP (Section~\ref{sec:additive}), whose inductive
bias tames the dimension.

\emph{Large $N$ (beyond a few thousand), any $D$.} The exact cubic computation is
infeasible; couple ARD with a scalable approximation
(Section~\ref{sec:scalable}). For general inputs, a stochastic variational GP
\citep{hensman2013} with an ARD kernel scales to very large $N$; for spatial or
moderate-dimensional problems the Vecchia-based method of \citet{cao2022} performs
simultaneous regression and selection at the scale of millions of observations.
The scalable spike-and-slab variational GP of \citet{dance2022} is the natural
choice when both scale and calibrated selection are required.

\emph{When the selection itself is the scientific object.} If the goal is not
prediction but a defensible statement about which inputs matter---with uncertainty,
and ideally with a guarantee---do not rely on raw lengthscales. Use a method built
for selection: spike-and-slab for inclusion probabilities, a hierarchical
selection prior for which consistency has been established
(Section~\ref{sec:theory}), or sensitivity and projection-predictive methods
\citep{piironen2016, paananen2019} that measure relevance by effect on prediction
rather than by rate of variation.

\emph{When linear and nonlinear effects must be compared.} Recall that ARD
overstates the relevance of nonlinear effects relative to linear ones of equal
predictive importance (Section~\ref{sec:disadvantages}); if the inputs mix the two,
prefer a sensitivity-based relevance measure
\citep{paananen2019} that ranks by contribution to the predictive distribution.

This guide is deliberately conservative: it recommends ARD wherever ARD is
trustworthy and escalates only where its specific limitations bite. That pattern
---ARD first, alternatives when warranted---is, in our view, the correct mental
model for the whole field, and it is the note on which we turn to applications and
then to broader discussion.

\section{Asymptotic Properties and Theoretical Guarantees}
\label{sec:theory}

A natural question for a statistical audience is whether ARD \emph{works} in the
asymptotic sense: as the sample size grows, does it recover the true set of
relevant inputs? The honest answer is that rigorous guarantees exist for several
relatives of ARD but not, in general, for plain lengthscale thresholding, and the
reasons for the gap are instructive. This section makes the theory precise. We
begin with the identifiability obstruction that fixed-domain asymptotics places on
the lengthscales themselves, then present the high-dimensional selection-
consistency theory available for hierarchical GP priors, reconcile the two
results---which concern different questions and are sometimes wrongly described as
paradoxical---and draw the implications for ARD.

\subsection{Identifiability under fixed-domain asymptotics}
\label{sec:identifiability}

The principal obstruction to estimating lengthscales is identifiability, and it
is sharpest under \emph{fixed-domain} (also called \emph{infill}) asymptotics, in
which observations accumulate with increasing density inside a bounded region.
\citet{zhang2004} showed that for the Mat\'ern family with fixed smoothness $\nu$,
not all covariance parameters can be estimated consistently: only a
\emph{microergodic} combination is. The mathematical mechanism is the equivalence
of Gaussian measures. Two zero-mean Gaussian measures on the space of functions
over a bounded domain are either mutually absolutely continuous (equivalent) or
mutually singular; when they are equivalent, no amount of data observed within the
domain can consistently distinguish their parameters, because the two models
assign comparable probability to every observable event.

The canonical instance is the exponential covariance $k(d) = \sigma^2
\exp(-d/\phi)$ with variance $\sigma^2$ and range $\phi$. \citet{zhang2004} proved
that under infill sampling the Gaussian measures induced by two parameter pairs
$(\sigma_1^2,\phi_1)$ and $(\sigma_2^2,\phi_2)$ are equivalent if and only if the
ratios match, $\sigma_1^2/\phi_1 = \sigma_2^2/\phi_2$; consequently only the
microergodic parameter $\sigma^2/\phi$ is consistently estimable and
asymptotically normal, while $\sigma^2$ and $\phi$ individually are not. Because
ARD reads the relevance of input $d$ off the individual lengthscale $\ell_d$
(the analogue of $\phi$ per coordinate), this is precisely the mathematical
reason that an unconditional consistency statement for the raw ARD lengthscales is
not to be expected: the data pin down a ratio, not the lengthscale itself, however
well the predictive distribution is determined. The relevance \emph{ranking} can
nonetheless be far more stable than the individual lengthscale \emph{values}, a
distinction we already drew in Section~\ref{sec:disadvantages} and that the
equivalence-of-measures result now explains.

\subsection{Selection consistency under high-dimensional asymptotics}
\label{sec:consistency}

The negative result of Section~\ref{sec:identifiability} concerns the
\emph{hyperparameters}. A different and more recent line of work shows that the
\emph{support}---which inputs are relevant---can be recovered consistently, under
a different asymptotic regime and for a model engineered to make selection
well-posed. \citet{jiang2021}, published in the \emph{Annals of Statistics},
establish variable-selection consistency for Gaussian process regression in a
\emph{high-dimensional} regime in which the predictor dimension $d_n$ is allowed
to grow with the sample size $n$.

It is important to be precise about the sampling model, because it is easy to
misdescribe. The inputs $X_1,\dots,X_n$ are taken to be independent draws from a
probability measure $Q_n$ on $\R^{d_n}$---in the main results, the $d_n$-variate
Gaussian $\Normal(0,\xi^2 I)$---and the predictor dimension $d_n$ grows with $n$.
This is high-dimensional asymptotics with i.i.d.\ random design; it is \emph{not}
the increasing-domain asymptotics of spatial statistics (in which a single random
field is observed over an expanding region), nor the fixed-domain regime of
Section~\ref{sec:identifiability}. The distinction matters because the two regimes
answer different questions, as we discuss in Section~\ref{sec:reconcile}.

The model assigns $f$ a rescaled Gaussian process prior hierarchically extended
with stochastic variable selection, and the truth $f_0$ is assumed to depend on a
sparse subset of $d_0$ relevant inputs. The main theorem shows that the posterior
probability of selecting the wrong subset vanishes,
\begin{equation}
\label{eq:consistency}
\E^{*}_n\!\left[\,\Pi_n\big(\Gamma \neq \gamma^{*}_n \,\big|\, \mathcal{D}_n\big)\right]
\;\longrightarrow\; 0
\qquad \text{as } n \to \infty,
\end{equation}
where $\Gamma$ is the random inclusion vector, $\gamma^{*}_n$ the true active set,
and $\E^{*}_n$ denotes expectation under the true data-generating distribution of
$\mathcal{D}_n$. The result holds at
least when $f_0$ has \emph{finite} smoothness, which is the condition that does
the work; we sketch why.

\begin{proofsketch}
The argument is a posterior-contraction argument in the empirical $L_2(Q_n)$
metric $\rho(f,f') = \{\int (f-f')^2 \, dQ_n\}^{1/2}$, in the tradition of Schwartz
theory and its rate-quantified descendants \citep{schwartz1965, ghosal2000}.
Three ingredients are needed. \emph{(i) Prior mass.} One verifies a
Kullback--Leibler prior-mass condition: the prior places enough mass on a
KL neighborhood of the truth,
\[
\Pi\!\left(f : \mathrm{KL}(p_{f_0}\| p_f) \le \epsilon_n^2,\ V(p_{f_0}\|p_f) \le \epsilon_n^2 \right)
\ \ge\ \exp(-n c\, \epsilon_n^2),
\]
which under the Gaussian design reduces to an $L_2(Q_n)$-ball statement about the
GP. \emph{(ii) Entropy and tests.} One constructs a sieve of functions whose
metric entropy is controlled, $\log N(\epsilon_n, \cdot, \|\cdot\|_{L_2(Q_n)}) \le
n\epsilon_n^2$, with the prior assigning exponentially small mass to its
complement, so that exponentially powerful tests separate the truth from
alternatives. \emph{(iii) Small-ball probabilities.} The technically demanding
step---and the paper's central contribution---is to bound the small-ball
probabilities of the rescaled GP \emph{at all rescaling levels}, which is achieved
by translating them, via the Gaussian design, into the metric entropy of the unit
ball of the reproducing-kernel Hilbert space associated with the kernel. The
finite smoothness $\beta$ of $f_0$ enters here: an over-specified model that
includes a spurious input enlarges the effective dimension of the function space,
and finite smoothness makes the associated prior-mass deficit \emph{polynomially}
(not merely logarithmically) large, producing a polynomial slowdown in the
contraction rate for false-positive models. The posterior therefore concentrates
on the true model, giving \eqref{eq:consistency}. The permissible dimension growth
is tied to the smoothness through a relation of the form $\log d_n \lesssim
n^{\,d_0/(2\beta+d_0)}$, so rougher true functions in fact permit larger candidate
dimensions.
\end{proofsketch}

The role of the finite-smoothness condition is worth emphasizing, because it is
the crux. The argument exploits exactly the same accounting as the Bayesian
Occam's razor of Section~\ref{sec:evidence}: including a spurious input buys
flexibility that the data do not reward, and the prior charges for it. Finite
smoothness is what makes the charge bounded away from zero; for an infinitely
smooth truth the rate depreciation for over-specified models is only logarithmic,
and the proof strategy does not deliver consistency. The marginal likelihood's
complexity penalty is, in these consistent hierarchical constructions, precisely
the finite-sample shadow of the prior-mass penalty that the contraction theory
exploits.

A complementary guarantee holds in the frequentist, penalized-likelihood setting.
For \emph{penalized-likelihood kriging}, \citet{chu2011} establish oracle-type
selection consistency for the mean component of a spatial linear model under
covariance tapering and a one-step sparse estimator. Here too the model has been
engineered---through an explicit penalty rather than an inclusion indicator---to
make selection a well-posed estimation problem.

\subsection{Reconciling the two results}
\label{sec:reconcile}

At first glance Sections~\ref{sec:identifiability} and~\ref{sec:consistency} seem
to collide: one says GP covariance parameters cannot be consistently estimated,
the other says GP variable selection is consistent. There is no contradiction, and
seeing why clarifies both results. They differ in two respects. First, they
concern \emph{different questions}: \citet{zhang2004} asks whether the covariance
\emph{hyperparameters} (variance, range) are identifiable, while \citet{jiang2021}
asks whether the \emph{support} (which inputs are relevant) is identifiable.
Non-identifiability of a continuous hyperparameter does not preclude recovery of a
discrete inclusion pattern; one can know which inputs matter without consistently
estimating how much. Second, they invoke \emph{different sampling models}: a single
spatial field densifying within a fixed domain (fixed-domain asymptotics) versus
$n$ independent covariate--response pairs in growing dimension (high-dimensional
asymptotics under i.i.d.\ design). We stress that the latter is not
``increasing-domain asymptotics,'' a distinct spatial-statistics framework in which
the observation region itself expands \citep{zhangzimmerman2005}; conflating the
two is a common slip. The reconciliation, then, is that the obstruction of
\citet{zhang2004} bites on hyperparameter estimation under infill sampling, while
the guarantee of \citet{jiang2021} concerns support recovery under a high-
dimensional design, and neither speaks to the other's regime.

\subsection{Implications for plain ARD}
\label{sec:theory_implications}

What does this theory say about ARD as practitioners actually use it---a single
product kernel, lengthscales fitted by Type-II maximum likelihood, relevance read
off the lengthscales, selection by a threshold? Less than one might hope, and the
gap is the honest headline. The consistency result of \citet{jiang2021} is proved
for a \emph{hierarchical GP prior with an explicit stochastic selection
mechanism}, not for marginal-likelihood ARD with lengthscale thresholding; the
consistent methods escape Zhang's obstruction not by estimating lengthscales
better but by changing the object of inference---to a discrete inclusion vector or
a penalized estimate---for which contraction can be established. Plain ARD makes no
such change: it continues to read relevance off the very lengthscales that infill
asymptotics shows are not consistently estimable. The practical reading is that
ARD's relevances inherit the predictive virtues of GPs without inheriting a
guarantee about the relevances themselves. They are an excellent heuristic and an
effective screening tool, but they should not be mistaken for consistent estimates
of a true active set; when a formal guarantee is required, one should adopt a
construction---hierarchical selection or penalized likelihood---for which such
guarantees have been proved. We regard the development of selection-consistency
theory directly for marginal-likelihood ARD, perhaps under design or smoothness
restrictions that restore identifiability, as an open and worthwhile problem, to
which we return in Section~\ref{sec:discussion}.

\section{Advantages, Disadvantages, and Failure Modes}
\label{sec:proscons}

Having explained how ARD works, we now weigh it. The method's popularity is
deserved, but it rests on a set of trade-offs that a careful user should
understand. We treat the advantages and the disadvantages in turn, and we return
to several of these threads quantitatively in the comparison of
Section~\ref{sec:compare} and computationally in Section~\ref{sec:comput}.

\subsection{Advantages}
\label{sec:advantages}

\paragraph{It is automatic and data-driven.} The defining virtue of ARD is in
its name. The analyst does not pre-specify which inputs matter, does not set a
relevance threshold a priori, and does not run a separate search over subsets.
Relevance is inferred jointly with the rest of the model by the same objective
that fits the function. In an emulation problem with dozens of inputs of unknown
importance, this is exactly the labor-saving automation one wants.

\paragraph{It is embedded in the standard GP workflow.} Turning ARD on requires
almost nothing. Every mainstream GP library exposes an ARD option that simply
replaces a scalar lengthscale with a vector---in GPy
\citep{gpy2014} one writes \texttt{RBF(input\_dim=D, ARD=True)}; GPflow
\citep{gpflow2017}, GPyTorch \citep{gardner2018}, GPML, and scikit-learn
\citep{pedregosa2011} all provide equivalents---and the same gradient-based
training routine is reused without modification. There is no new inference
algorithm to implement, debug, or tune. This near-zero marginal cost of adoption
is, we will argue in Section~\ref{sec:whypopular}, the single most important
reason ARD dominates in practice.

\paragraph{It inherits the GP's uncertainty quantification.} Because ARD lives
inside a GP, it comes with the posterior predictive distribution
\eqref{eq:posterior} for free. Predictions carry calibrated error bars, and the
relevance estimates themselves can be equipped with uncertainty in the fully
Bayesian treatment (Section~\ref{sec:fullbayes}). For applications in which
honest uncertainty is the whole point---calibration of expensive simulators,
sequential design, safety-critical prediction---this is a decisive advantage over
selection methods that return only a point estimate of which variables matter.

\paragraph{It is nonparametric and flexible.} ARD does not assume linearity,
additivity, or any particular functional form. It captures nonlinear effects and,
through the product structure of the kernel, interactions among inputs. A
variable can be selected as relevant because it matters nonlinearly or only in
concert with others, situations that coefficient-based linear selection would
miss entirely.

\paragraph{Its output is interpretable.} A vector of lengthscales is easy to
read: sort the inputs by $1/\hat\ell_j$ and the most relevant rise to the top.
This transparency is valuable in scientific settings where the selection is not
merely a means to better prediction but an end in itself---an answer to the
question of which factors drive the response.

\subsection{Disadvantages}
\label{sec:disadvantages}

\paragraph{The optimization is non-convex and multimodal.} As discussed in
Section~\ref{sec:landscape}, the marginal likelihood may have multiple local
optima, and gradient ascent can converge to a poor one. Different
initializations can yield different relevance patterns. In low dimension this is
usually managed by a handful of random restarts, but the problem worsens as $D$
grows and the surface acquires more basins. The relevances ARD reports are
conditional on the optimizer having found a good mode, an assumption that is
rarely verified and occasionally false.

\paragraph{It is computationally expensive.} Each evaluation of
\eqref{eq:logml} and its gradient requires factorizing the $N \times N$ matrix
$\Ky$, an $O(N^3)$ operation with $O(N^2)$ storage, and this must be repeated at
every step of the optimization. For $N$ beyond a few thousand the cost becomes
prohibitive without approximation. ARD compounds the standard GP burden because
it adds $D$ hyperparameters, enlarging the optimization and, as we discuss next,
making each restart more likely to be needed. Section~\ref{sec:comput} is
devoted to these issues and their remedies.

\paragraph{It struggles as the input dimension grows.} The number of lengthscale
hyperparameters equals the input dimension, so the optimization problem inflates
with $D$. Worse, the very thresholding step that converts lengthscales into
selections becomes unreliable in high dimension: with many inputs, the
relevances of truly irrelevant variables spread out and can overlap the
relevances of weakly relevant ones, blurring the boundary that
Figure~\ref{fig:relevance} drew so cleanly in a low-dimensional example. This
unreliability of lengthscale thresholding in high dimension is precisely the
motivation given by \citet{dance2022} for replacing it with a probabilistic
spike-and-slab inclusion model.

Part of the difficulty is geometric. The squared-exponential kernel acts on the
weighted distance between inputs, and in high dimension distances concentrate: for
many distributions, as $D \to \infty$ the pairwise Euclidean distances between
i.i.d.\ points become increasingly similar to one another, so the ratio of the
nearest to the farthest distance tends to one. When all pairwise distances are
nearly equal, the off-diagonal kernel entries become nearly constant, and a
covariance dominated by a constant plus noise carries little information to
distinguish a relevant direction from an irrelevant one---in the limiting picture
the kernel matrix approaches a scaled identity, the marginal-likelihood gradients
along individual lengthscales flatten, and the optimizer has correspondingly weak
signal with which to inflate the irrelevant lengthscales. This is one mechanism
behind the eroding margin we now illustrate, and it is intrinsic to distance-based
kernels rather than a defect of the optimizer.

\begin{figure}[t]
\centering
\includegraphics[width=0.60\textwidth]{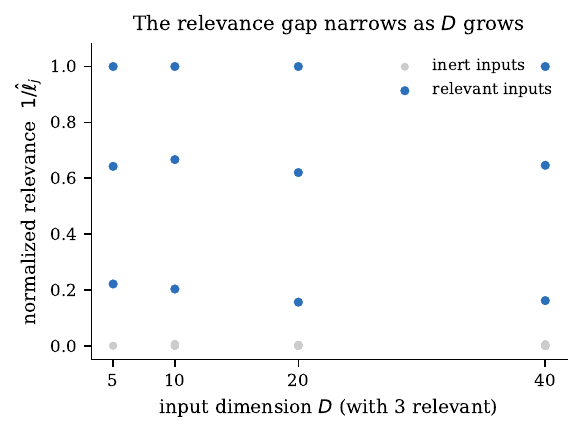}
\caption{Relevances of three fixed relevant inputs (filled) and the surrounding
inert inputs (grey) as the input dimension $D$ grows, each from a refitted
SE-ARD GP with $N=150$. The two strongest relevant inputs stay well above the
inert floor, but the weakest drifts toward it as $D$ increases, eroding the
margin a thresholding rule relies on. Relevances are normalized by the maximum
within each fit.}
\label{fig:highdim}
\end{figure}

Figure~\ref{fig:highdim} illustrates the phenomenon concretely. We fixed three
relevant inputs of decreasing strength and embedded them among an increasing
number of inert inputs, refitting an SE-ARD GP at each dimension. The two
strongest relevant inputs remain clearly separated from the inert floor even at
$D=40$, which is a genuine and reassuring success of ARD; but the
\emph{weakest} relevant input drifts steadily downward toward the inert cluster
as $D$ grows, and by $D=40$ the gap that would let a thresholding rule cleanly
separate it from the noise has nearly closed. The lesson is not that ARD
collapses in high dimension---it degrades gracefully rather than catastrophically
---but that the margin protecting weak signals erodes, so that the inputs most
in need of careful selection are exactly the ones the threshold is least able to
resolve. This is the regime in which the calibrated inclusion probabilities of a
spike-and-slab model \citep{dance2022} or the strong shrinkage of a SAAS prior
\citep{eriksson2021} repay their additional cost.

\paragraph{The lengthscales are only weakly identified.} A subtle but important
limitation concerns identifiability. Under fixed-domain (infill) asymptotics,
\citet{zhang2004} showed that for the Mat\'ern class not all covariance
parameters can be estimated consistently: only certain combinations---the
so-called microergodic parameters---are pinned down by the data, while the
individual range parameters and the variance are confounded and only their
ratio is well determined. Translated to ARD, this means the individual
lengthscales can have high estimation variance even when the predictive fit is
excellent. The relevance \emph{ranking} is often more stable than the individual
lengthscale \emph{values}, which is a reason to treat the magnitudes with
caution. The practical symptom is familiar to anyone who has fitted GPs: two runs
that achieve nearly identical marginal likelihoods and nearly identical
predictions can report visibly different lengthscales, because the data constrain
the predictive distribution far more tightly than they constrain the individual
relevance parameters. For variable selection this is mostly benign when the
question is the ranking---which is comparatively robust---and more troubling when
the question is the magnitude, for instance when one wishes to report \emph{how
much} more relevant one input is than another. A prudent practice is to report
relevances on a relative scale and to avoid over-interpreting small differences,
and, when the magnitudes genuinely matter, to propagate hyperparameter uncertainty
through the fully Bayesian treatment of Section~\ref{sec:fullbayes} rather than
trusting a single point estimate.

\paragraph{It can misjudge relevance across effect types.} A counterintuitive
and underappreciated failure mode was documented by \citet{piironen2016} and
\citet{paananen2019}: ARD systematically overstates the relevance of inputs that
enter \emph{nonlinearly} relative to inputs that enter \emph{linearly}, even when
the two contribute equally to the response in a mean-squared-error sense. The
reason is that the lengthscale measures rate of variation, not contribution to
predictive accuracy; a strongly curved but small-amplitude effect can earn a
short lengthscale while a large-amplitude linear effect earns a long one. An
analyst who reads $1/\hat\ell_j$ as ``importance for prediction'' can therefore be
misled, and this is a principal motivation for the sensitivity-based and
projection-based alternatives of Section~\ref{sec:sensitivity}, as well as for
additive methods that classify each effect as zero, linear, or nonlinear outright
(Section~\ref{sec:additive}).

\paragraph{It delivers only effective sparsity, and can overfit the evidence.} As
emphasized in Section~\ref{sec:effectivesparsity}, ARD ranks rather than selects,
and the lack of exact zeros means a thresholding rule is always required.
Moreover, optimizing a high-dimensional hyperparameter vector by maximizing the
evidence can itself overfit: \citet{qi2004} showed that evidence maximization
can drive lengthscales to extreme values that improve the marginal likelihood but
degrade predictive performance, and proposed a predictive variant of ARD that
monitors leave-one-out predictive accuracy instead. The phenomenon is the
hyperparameter-level analogue of overfitting and is more pronounced when $D$ is
large relative to $N$.

\paragraph{It carries no general guarantee of selecting the right variables.}
Finally, and despite its excellent empirical record, plain ARD with lengthscale
thresholding comes with no general theorem certifying that it recovers the true
active set as $N$ grows. Selection-consistency results do exist for related
constructions---hierarchical GP priors with explicit selection indicators, and
penalized-likelihood kriging---but they do not cover lengthscale thresholding
per se, and the identifiability obstruction of \citet{zhang2004} suggests why a
clean consistency statement for the raw lengthscales is hard to come by. We
review what is known in Section~\ref{sec:theory}.

\subsection{Summary of the trade-off}

The picture that emerges is coherent. ARD is the method of choice when the input
dimension is low to moderate, the sample size is small enough for exact GP
computation (or amenable to the approximations of Section~\ref{sec:comput}), and
the analyst wants a quick, interpretable, uncertainty-aware ranking of inputs
with essentially no extra implementation effort. It is the method to approach
with caution---or to replace with one of the alternatives of
Section~\ref{sec:compare}---when the dimension is high, when exact zeros or
calibrated inclusion probabilities are required, when linear and nonlinear
effects must be compared on an equal footing, or when a formal selection
guarantee matters. The next section makes these comparisons concrete.

\section{The Frontier Landscape of GP Variable Selection}
\label{sec:compare}

ARD is one point in a rich space of methods for deciding which inputs a Gaussian
process should depend on. This section surveys that space, organizing the methods
by the strategy they use to express and discover relevance, and then returns to
the question that motivates the whole comparison: given so many principled
alternatives, why does ARD remain the default? Table~\ref{tab:compare} summarizes
the comparison along the axes that matter most in practice.

\subsection{Probabilistic exact-sparsity methods}
\label{sec:exactsparsity}

The first family of alternatives confronts ARD's defining limitation---
effective rather than exact sparsity (Section~\ref{sec:effectivesparsity})---
head on, by building genuine sparsity into the prior and returning calibrated
probabilistic statements about inclusion rather than point relevances.

\subsubsection{Spike-and-slab priors on relevance}
\label{sec:spikeslab}

The most direct fix for ARD's lack of exact zeros is to build the zeros into the
prior. Spike-and-slab variable selection, a mainstay of Bayesian linear
modeling \citep{george1993}, attaches to each input a binary inclusion indicator
$\gamma_j \in \{0,1\}$: when $\gamma_j = 0$ the input is excluded (its relevance
is exactly zero, the ``spike''), and when $\gamma_j = 1$ the input is active and
its relevance is drawn from a continuous ``slab.'' Posterior inference over
$\bgamma$ then yields, for each input, a \emph{posterior inclusion
probability}---a calibrated, probabilistic answer to ``is this variable
relevant?'' that ARD's point lengthscales cannot provide.

For Gaussian processes, \citet{linkletter2006} placed such indicators on the
correlation parameters of a computer-experiment GP and, in a contribution that
has outlived its original setting, introduced the device of augmenting the design
with a known \emph{inert} input. Because this artificial variable is guaranteed
irrelevant, its estimated importance furnishes a reference distribution---a noise
floor---against which the importance of a genuine input is judged. This is the
calibration idea we used in Figure~\ref{fig:relevance}, and it can be grafted onto
ARD itself. \citet{savitsky2011}, in a paper that we recommend as a model of
exposition on exactly this topic, developed a full
spike-and-slab framework for nonparametric GP priors, with Bernoulli inclusion
indicators on the inputs and Markov chain Monte Carlo for inference, and compared
the resulting models across several computational strategies.

The historical cost of spike-and-slab GPs was computational: marginalizing or
sampling the discrete indicators on top of the cubic GP cost was expensive.
Recent work has largely removed this barrier. \citet{dance2022} give a scalable
variational spike-and-slab GP that produces posterior inclusion probabilities at
a cost comparable to ordinary sparse variational GPs, and they motivate their
method precisely by the observation that thresholding ARD lengthscales is
unreliable in high dimension. Spike-and-slab is thus the natural upgrade from ARD
when one needs exact sparsity or calibrated inclusion probabilities and can afford
a modest increase in machinery.

It is illuminating to see how spike-and-slab relates to ARD formally. One way to
write the connection is to parameterize the kernel with inverse squared
lengthscales $\rho_j = 1/\ell_j^2$, so that $\rho_j = 0$ switches input $j$ off
exactly. Plain ARD places no prior on $\rho_j$ and estimates it by maximum
marginal likelihood, which yields $\hat\rho_j$ small but positive for irrelevant
inputs. A spike-and-slab model instead places on each relevance a two-point
mixture of a point mass at zero (the spike) and a continuous slab,
\begin{equation}
\label{eq:spikeslab}
\rho_j \mid \gamma_j \;\sim\; (1-\gamma_j)\,\delta_0 \;+\; \gamma_j\,g_{\mathrm{slab}},
\qquad
\gamma_j \sim \mathrm{Bernoulli}(\pi),
\end{equation}
equivalently $\rho_j = \gamma_j\,\tilde\rho_j$ with $\tilde\rho_j \sim
g_{\mathrm{slab}}$ a continuous slab density and $\delta_0$ the Dirac mass at
zero; integrating over $\gamma_j$ produces a posterior that places genuine mass at
$\rho_j = 0$ and a posterior inclusion probability $\Pr(\gamma_j = 1 \mid
\mathcal{D})$ for each input. Inference is no longer a continuous optimization but
a trans-dimensional sampling problem over the $2^D$ models indexed by
$\bgamma \in \{0,1\}^D$, a genuine paradigm shift in computation. The fully
Bayesian SAAS prior of Section~\ref{sec:fullbayes} can be seen as a continuous
relaxation of \eqref{eq:spikeslab}, replacing the discrete spike with a
heavy-tailed continuous prior that concentrates near zero. Viewed this way, ARD,
spike-and-slab, and SAAS form a progression in how aggressively and how honestly
the model expresses the belief that most inputs are irrelevant---from a bare point
estimate, through a continuous shrinkage prior, to an explicit discrete
mixture---with computational cost and selection fidelity both increasing along the
progression.

\subsubsection{Fully Bayesian relevance and the SAAS prior}
\label{sec:fullbayes}

A second response to ARD's weaknesses keeps the continuous lengthscale
parameterization but abandons the point estimate. Instead of maximizing the
marginal likelihood over $\btheta$, one places a prior on the lengthscales and
\emph{integrates} over them, propagating hyperparameter uncertainty into the
predictions and the relevance assessment. Because the resulting posterior is not
available in closed form, inference proceeds by Markov chain Monte Carlo---
Hamiltonian Monte Carlo and its No-U-Turn variant \citep{neal2011,hoffman2014},
or slice sampling tailored to GP hyperparameters \citep{murray2010}. The payoff
is twofold: the multimodality that traps a point optimizer
(Section~\ref{sec:landscape}) is explored rather than collapsed, and relevance
comes with a posterior distribution rather than a single number.

A prominent recent instance, developed for high-dimensional Bayesian
optimization but broadly applicable, is the sparse axis-aligned subspace (SAAS)
prior of \citet{eriksson2021}. Writing $\rho_d = 1/\ell_d^2$ for the inverse
squared lengthscale of input $d$, the SAAS prior is the hierarchy
\begin{equation}
\label{eq:saas}
\tau \sim \mathcal{HC}(\alpha),
\qquad
\rho_d \mid \tau \overset{\mathrm{iid}}{\sim} \mathcal{HC}(\tau),
\quad d = 1,\dots,D,
\end{equation}
where $\mathcal{HC}$ denotes the half-Cauchy distribution and $\tau$ is a global
shrinkage parameter. The structure is exactly that of a global--local shrinkage
prior: the global $\tau$ pulls all inverse lengthscales toward zero (switching
inputs off), while the heavy Cauchy tails on each $\rho_d$ let a few coordinates
escape the shrinkage and take large values (switching those inputs on). This is
the continuous, differentiable mechanism by which SAAS approximates the behavior
of an exact $\ell_0$ penalty while remaining amenable to gradient-based Hamiltonian
sampling. Inference is by the No-U-Turn sampler. The authors report that sampling,
rather than maximizing, is essential for recovering sparse structure in high
dimension---a direct empirical indictment of point-estimate ARD in that regime,
and a demonstration that the continuous parameterization can be made to select well
when paired with a strong sparsity prior and honest integration.

The cost, of course, is computation: MCMC over GP hyperparameters is far more
expensive than a few gradient steps, and each sampler iteration still pays the
cubic price of a covariance factorization. This is the central trade that
Table~\ref{tab:compare} records---principled uncertainty and better
high-dimensional selection, bought with substantially more computation.

\subsubsection{Global--local shrinkage: the Laplace and horseshoe priors}
\label{sec:globallocal}

The SAAS prior \eqref{eq:saas} is worth a second look, because its structure
connects ARD to a large and theoretically mature literature on \emph{global--local
shrinkage} priors that the GP community has only partly absorbed
\citep{polsonscott2011}. A global--local prior writes each relevance as a product
of a global scale, shared across all inputs and pulling everything toward zero,
and a local scale, specific to each input and able to rescue it from the
shrinkage. The art is entirely in the choice of the local prior, and two choices
anchor the spectrum.

The \emph{Laplace} prior---the Bayesian counterpart of the lasso
\citep{parkcasella2008}---takes each relevance to have a double-exponential
distribution, representable as a Gaussian scale mixture with an exponential
mixing density. This is the prior whose posterior \emph{mode} is the lasso
estimate, and it inherits the lasso's defining behavior and its defining flaw. A
single global penalty controls all coordinates, and because the double-exponential
has \emph{exponential} (light) tails, that one penalty cannot simultaneously leave
genuinely large relevances unshrunk and drive small ones to zero: tuning it
aggressively enough to kill noise also biases the true signals, while tuning it
gently enough to spare the signals lets noise leak through. Moreover, although the
posterior mode is sparse, the posterior \emph{mean}---the natural Bayesian point
estimate---is not \citep{parkcasella2008, castillo2015}. For variable selection
this is the same effective-sparsity caveat that afflicts ARD
(Section~\ref{sec:effectivesparsity}), now in Bayesian dress.

The \emph{horseshoe} prior \citep{carvalho2009, carvalho2010} repairs exactly this
defect by giving each local scale a half-Cauchy distribution. The half-Cauchy has
two features the Laplace lacks: an infinite spike at the origin, which shrinks
noise coordinates hard, and heavy, Cauchy-like tails, which let a few large
relevances escape the global shrinkage essentially untouched. The result behaves
much more like a spike-and-slab prior than a single-scale penalty does---strong
shrinkage of the many, near-zero shrinkage of the few---and it comes with
supporting theory: for the sparse normal-means problem, the horseshoe posterior
attains the minimax estimation rate (up to a constant, when the number of nonzero
signals is known or estimated) and is provably more informative than one-component
priors such as the Laplace \citep{vanderpas2014}. The regularized or ``Finnish''
horseshoe \citep{piironen2017} tempers the tails to improve stability in practice.
\citet{bhadra2019} survey the contrast between the convex,
fast, gold-standard lasso and the non-convex, state-of-the-art horseshoe in full.
Global--local priors of this kind are by now a workhorse for high-dimensional
variable selection across regression settings: \citet{ma2025}, for example, build
a hierarchical model with horseshoe and normal-gamma priors for selection in
binary quantile regression, and scalable computation for the associated
sparsity-inducing priors continues to advance \citep{paul2025}.

The relevance of all this to ARD is that the SAAS prior of
Section~\ref{sec:fullbayes} is, structurally, a horseshoe-type prior placed on the
GP inverse squared lengthscales: a half-Cauchy global scale $\tau$ multiplying
half-Cauchy local scales $\rho_d$ in \eqref{eq:saas} is precisely the global--local
hierarchy that defines the horseshoe. (The one nuance is that the classical
horseshoe places this hierarchy on the \emph{scale of a Gaussian-distributed
coefficient}, whereas SAAS places it on the kernel \emph{relevance} parameters
themselves; SAAS is thus horseshoe-type in its hyperprior, not literally the
normal-means horseshoe estimator.) Seen this way, the progression from ARD to
SAAS is the same progression that the shrinkage literature traces from the lasso
to the horseshoe: from a point estimate of relevance, through a single-scale
Laplace penalty, to a heavy-tailed global--local prior that shrinks noise and
spares signal at once. The connection is more than analogy. \citet{castillo2025}
place a horseshoe prior directly on the GP lengthscales and prove that the
resulting process attains near-minimax, dimension-adaptive posterior contraction
rates while performing a form of variable selection by ``freezing'' the dimensions
whose lengthscales vanish; they also show that an \emph{exponential}
(Laplace-type) lengthscale prior is, by contrast, dimension-suboptimal in the
high-dimensional regime---the GP analogue of the horseshoe-beats-lasso phenomenon.
The practical lesson for the ARD user is that if point-estimate relevances are not
enough, the most principled upgrade is not a single-scale Bayesian penalty but a
global--local one, and that the half-Cauchy SAAS prior is already an instance of
the right idea.
\label{sec:penalizedgroup}

A frequentist route to exact sparsity replaces the implicit Occam penalty of
the marginal likelihood with an explicit, sparsity-inducing penalty, importing
the lasso's machinery into the kernel.

\subsubsection{Penalized-likelihood kriging}
\label{sec:penalized}

The final family imports the lasso's machinery directly. Rather than relying on
the Occam's-razor penalty implicit in the marginal likelihood, one adds an
\emph{explicit} sparsity-inducing penalty on the relevance parameters and
optimizes the penalized likelihood, obtaining exact zeros as in the lasso. The
appeal is precisely that it cures ARD's central deficiency---effective rather
than exact sparsity (Section~\ref{sec:effectivesparsity})---by borrowing the
mechanism that makes the lasso \citep{tibshirani1996} set coefficients exactly to
zero. \citet{li2005} applied this idea to Gaussian kriging models for computer
experiments, penalizing the correlation parameters to perform selection;
\citet{yi2011} developed penalized GP regression and classification for
high-dimensional nonlinear data with lasso- and SCAD-type penalties on the inverse
lengthscales, demonstrating that the approach scales to the regime where the
number of candidate inputs is large. On the spatial-statistics side,
\citet{chu2011} combined covariance tapering---which sparsifies the covariance
matrix and so addresses the computational bottleneck at the same time---with a
one-step sparse estimator, and, importantly for Section~\ref{sec:theory},
established selection consistency for the mean component of a spatial linear
model. The same penalize-the-kernel idea appears beyond kriging: \citet{zhao2025}
construct a doubly sparse garrotized kernel machine that performs simultaneous
variable selection and estimation in a high-dimensional partially linear model,
illustrating how naturally sparsity penalties graft onto kernel methods more
broadly. Penalized-likelihood methods are the natural choice when exact sparsity is
wanted within a frequentist, point-estimation workflow, and when one is willing to
set the penalty's tuning parameter by cross-validation or an information
criterion. Their relationship to ARD is closer than it may appear: as
\citet{wipf2008} showed, the ARD objective itself can be rewritten as a sequence
of reweighted $\ell_1$ problems, so penalized kriging can be seen as making
explicit, and exact, the shrinkage that ARD performs implicitly and softly.

The penalty idea also has a fully Bayesian counterpart that blurs the line between
this family and the probabilistic priors of
Section~\ref{sec:exactsparsity}. Recent work on Bayesian bridge GP regression
\citep{b2gpr2025} places $\ell_q$-norm constraints directly on the GP parameters,
recovering a conjugate Gaussian prior at $q=2$ and, for $0<q<2$, a constrained
flat prior whose norm-constrained posterior induces sparsity in the manner of
bridge regression; inference uses a Gibbs sampler with a spherical Hamiltonian
Monte Carlo step to sample efficiently from the constrained posterior. Such
constructions show that the explicit-penalty and the prior-based routes to
sparsity are two views of a single underlying idea---a regularizer on the
relevance parameters---differing mainly in whether it is optimized or integrated.

\subsection{Dependence, sensitivity, and projection methods}
\label{sec:sensitivitygroup}

A third family steps outside the GP's own likelihood and measures relevance by
a criterion of statistical dependence, predictive sensitivity, or projection onto
a low-dimensional structure.

\subsubsection{Kernel-dependence and sensitivity criteria}
\label{sec:sensitivity}

A third family steps outside the GP's own likelihood and measures relevance by a
criterion of statistical \emph{dependence} or predictive \emph{sensitivity}.

The Hilbert--Schmidt independence criterion (HSIC) of \citet{gretton2005} is a
kernel-based measure of dependence between an input and the response that is zero
if and only if they are independent (for characteristic kernels). Using HSIC as
the objective, \citet{song2012} built backward-elimination feature selectors that
greedily remove the input contributing least to the dependence. These methods are
model-agnostic---they do not require a GP at all---but they sit naturally
alongside GP regression and provide an alternative notion of relevance grounded
in dependence rather than in lengthscale.

Closer to the GP, \citet{piironen2016} and \citet{paananen2019} rank inputs by
the \emph{sensitivity of the posterior predictive distribution}: an input is
relevant to the extent that perturbing it changes the GP's predictions. A
concrete and widely used score is the expected squared magnitude of the
predictive mean's partial derivative with respect to input $j$,
\begin{equation}
\label{eq:sensitivity}
S_j \;=\; \E_{\bx}\!\left[\left(\frac{\partial \bar f(\bx)}{\partial x_j}\right)^{\!2}\right],
\end{equation}
the expectation taken over the input distribution; an alternative measures the
Kullback--Leibler divergence of the predictive distribution under a small
perturbation of $x_j$. Because the derivative of a GP is itself a GP, the
derivative process and hence \eqref{eq:sensitivity} are available in closed form,
and a related thread \citep{blix2018} scores inputs directly through posterior
partial derivatives. The signal contribution of this line of work is conceptual:
unlike the lengthscale, the score \eqref{eq:sensitivity} measures relevance by
\emph{effect on prediction}, which is precisely the quantity that ARD's
lengthscales fail to track when linear and nonlinear effects are mixed
(Section~\ref{sec:disadvantages}). A short lengthscale signals rapid variation,
but rapid variation of small amplitude contributes little to prediction, whereas a
gentle but large-amplitude trend contributes much; the integrated derivative
\eqref{eq:sensitivity} does not confuse the two. The projection-predictive
approach of \citet{piironen2016} goes further, projecting a rich reference GP onto
sparse submodels and selecting the smallest submodel that predicts almost as
well---importing into the GP world the decision-theoretic selection philosophy
that has proved successful for generalized linear models, in which one first fits
the best available predictive model and then asks how much of its predictive
performance a sparse approximation can retain. The cost of these methods is
moderate---they require a fitted reference GP and then a search over submodels or a
computation of sensitivities---but they directly target ARD's
linear-versus-nonlinear blind spot, and for that reason they are the natural choice
when the relevance ranking, rather than the prediction, is what the analysis is
for.

\subsubsection{Active subspaces and linear projection}
\label{sec:activesubspace}

The methods so far select among the original coordinates. Sometimes the function
depends not on individual inputs but on a few \emph{linear combinations} of them,
and the right notion of dimension reduction is a subspace rather than a subset.
The active-subspace methodology of \citet{constantine2014,constantine2015} finds
such directions from the eigenvectors of the average outer product of the
gradient, $C = \E[\nabla f \, \nabla f^\top]$; the dominant eigenvectors span the
directions along which the function varies most on average, and a gap in the
eigenvalue spectrum identifies a low-dimensional active subspace onto which the
function can be projected with little loss. When $f$ is replaced by a GP
surrogate, both the matrix $C$ and its eigendecomposition can be computed in
closed form---because the gradient of a GP is itself a GP, the required
expectation is available analytically---marrying active subspaces to GP emulation
and making the whole construction a post-processing step on a fitted GP. The
connection to ARD is through the metric $M$ in the kernel \eqref{eq:seard}:
diagonal $M$ recovers coordinate ARD, in which the eigenvectors are forced to
align with the coordinate axes, while a general symmetric positive-definite $M$
recovers a linear-projection model whose leading eigenvectors play the role of an
active subspace \citep[][\S5.1]{rw2006}. Active subspaces are therefore the method
of choice when the relevant structure is genuinely a rotation of the input
space---when, say, the response depends on $x_1 + x_2$ but not on $x_1 - x_2$---a
situation in which coordinate-wise ARD is mis-specified and will report both
$x_1$ and $x_2$ as relevant without discovering that only their sum matters. The
price is that a subspace is less interpretable than a subset: a direction that
mixes many inputs does not answer the scientific question ``which variables
matter?'' as cleanly as a list does, and the choice between coordinate selection
and subspace projection ultimately turns on whether the inputs have individual
meaning that a rotation would destroy. A construction tailored to GP regression in
this spirit is the sparse projection of \citet{park2022}, which learns a
low-dimensional projection of the inputs jointly with the GP and encourages the
projection matrix to be sparse, so that the recovered directions involve few
original variables and thereby retain some of the interpretability that a dense
rotation sacrifices---a useful middle ground between coordinate ARD and a fully
general active subspace.

\subsection{Structured additive kernels}
\label{sec:additivegroup}

A fourth family abandons the single product kernel for a structured sum, trading
a flat relevance ranking for an interpretable decomposition of the function.

\subsubsection{Additive and compositional kernels}
\label{sec:additive}

A different philosophy abandons the single product kernel in favor of a
\emph{structured sum}. \citet{duvenaud2011} introduced additive Gaussian
processes, in which the kernel is a sum over interaction orders---first-order
terms in each input, second-order terms in each pair, and so on---each built from
one-dimensional base kernels. Concretely, writing $k_j$ for a base kernel acting
on input $j$ alone, the first-order additive kernel is $\sum_j k_j$, the
second-order kernel adds $\sum_{j<j'} k_j k_{j'}$, and the full additive kernel
sums all orders, each weighted by an order-variance hyperparameter. The
decomposition is interpretable in a way the monolithic ARD kernel is not: one can
read off how much of the function's variance is explained by each input
individually, by each pairwise interaction, and so forth, turning variable
selection into a question about which terms in the expansion carry weight. The
order variances themselves act as relevances at the level of interaction order,
and the per-input base-kernel lengthscales can carry ARD within each term, so the
additive model nests ARD rather than replacing it. \citet{duvenaud2013} extended
the idea to an automatic search over kernel \emph{structure}, composing kernels
by addition and multiplication to discover the form of the function, including
which inputs and interactions to include; this is variable selection and model
selection performed jointly, at the cost of a search over a combinatorial space
of kernel expressions.

Additive models are especially attractive when interactions are of scientific
interest, or when the analyst wants a decomposition of variance rather than a
flat list of relevances. Their cost is a larger model space and, in the
structure-search version, a combinatorial search; their relationship to ARD is
complementary rather than competitive, and the two are sometimes combined, with
ARD lengthscales inside each additive component. A further benefit, noted by
several authors, is statistical: low-order additive models impose a strong
inductive bias that mitigates the curse of dimensionality, so that an additive GP
can sometimes select and predict well in dimensions where a full product-kernel
ARD model struggles.

The additive viewpoint also offers a clean route around the linear-versus-
nonlinear blind spot of Section~\ref{sec:disadvantages}. The recently proposed
variational automatic relevance determination (VARD) method of \citet{vard2026}
fits a sparse additive regression model in which each feature's one-dimensional
component is estimated by a variational ARD procedure, and---crucially---the
method classifies every feature's contribution as exactly \emph{zero},
\emph{linear}, or \emph{nonlinear}, rather than collapsing relevance onto a single
lengthscale. Although developed for additive models rather than full GP regression,
VARD shares ARD's evidence-based philosophy while directly delivering the
distinction that raw GP lengthscales obscure, and it points to a broader lesson:
when the effect type matters, building the linear/nonlinear split into the model,
as the additive decomposition does, is more reliable than trying to read it off a
product kernel after the fact.

\subsection{Why ARD remains popular}
\label{sec:whypopular}

\begin{table}[t]
\centering
\caption{ARD compared with the principal alternatives for Gaussian process
variable selection. ``UQ'' denotes uncertainty quantification for the selection
itself; ``PIP'' denotes posterior inclusion probabilities. Computational cost is
relative and refers to the selection machinery on top of the base GP.}
\label{tab:compare}
\begin{tabular}{@{}lccccc@{}}
\toprule
Method & Sparsity & Selection UQ & Rel.\ cost & Scales in $D$ & Software \\
\midrule
ARD (Type-II ML)            & effective & none (point)   & low       & moderate & ubiquitous \\
Spike-and-slab GP           & exact     & PIP            & high\,$^{a}$ & good     & some \\
Fully Bayesian / SAAS       & effective & full posterior & high      & good     & some \\
Global--local (horseshoe)   & effective & full posterior & high      & good     & limited \\
Additive / compositional    & per-term  & partial        & moderate  & moderate & some \\
HSIC / dependence           & subset    & none           & low       & good     & some \\
Sensitivity / projection    & ranking   & partial        & moderate  & good     & limited \\
Active subspaces            & subspace  & none           & moderate  & good     & limited \\
Penalized kriging           & exact     & none           & moderate  & good     & limited \\
\bottomrule
\end{tabular}

\vspace{2pt}
{\footnotesize $^{a}$ Reduced to near-ARD cost by recent variational methods
\citep{dance2022}.}
\end{table}

The alternatives just surveyed are, in various respects, more principled than
ARD: spike-and-slab gives exact sparsity and inclusion probabilities, fully
Bayesian inference honestly propagates uncertainty, sensitivity methods measure
the right notion of relevance, and penalized kriging comes with consistency
theory. Why, then, does ARD remain the method most practitioners reach for first?

The reasons are pragmatic, and Table~\ref{tab:compare} makes them visible. ARD
requires \emph{no new inference machinery}: it is the anisotropic default of the
same kernel one would use anyway, and it is fitted by the same marginal-likelihood
optimization that one would run regardless. It is \emph{available everywhere}---a
single Boolean flag in every major library \citep{gpy2014, gpflow2017,
gardner2018, pedregosa2011}---so the cost of trying it is essentially zero. It is
\emph{cheap relative to the fully Bayesian and spike-and-slab alternatives},
needing a few gradient-based optimizations rather than thousands of MCMC
iterations. Its output is \emph{immediately interpretable} as a relevance
ranking. And in the regime where most GP modeling actually happens---low to
moderate input dimension, modest sample size---it \emph{performs well
empirically}, recovering the relevant inputs in cases like
Figures~\ref{fig:lengthscale} and~\ref{fig:relevance} without fuss. ARD is, in
short, the default for the same reason that many defaults persist: it is good
enough, almost free, and already implemented. The alternatives earn their keep
precisely where ARD's limitations bite---high dimension, the need for exact zeros
or calibrated probabilities, mixed linear and nonlinear effects, or a demand for
formal guarantees---and a sophisticated practitioner treats the choice as
contingent on which of those features the problem at hand requires.

\section{Scalability and Large-$N$, Large-$p$ Regimes}
\label{sec:comput}

ARD inherits the computational burden of Gaussian processes and adds difficulties
of its own. This section anatomizes those difficulties and surveys the
approximations that have made ARD usable at scale; following the usual ``large
$n$, large $p$'' idiom, we write $p$ for the input dimension (denoted $D$
elsewhere) when emphasizing the high-dimensional regime. The material here is what
separates a method that works on a textbook example from one that works on a real
data set.

\subsection{The cubic bottleneck}
\label{sec:cubic}

The dominant cost of GP regression is the solution of linear systems involving
the $N \times N$ matrix $\Ky = \Kff + \sigma_n^2 I$. Both the predictive
equations \eqref{eq:postmean}--\eqref{eq:postvar} and the marginal likelihood
\eqref{eq:logml} require $\Ky^{-1}$ or, in practice, a Cholesky factorization
$\Ky = LL^\top$. Computing this factorization costs $O(N^3)$ floating-point
operations and $O(N^2)$ memory \citep[][\S2.3]{rw2006}. Once $L$ is available,
each subsequent solve is $O(N^2)$, but the factorization itself is the wall: at
$N = 10^4$ it is already a second or more per evaluation, and at $N = 10^5$ a
dense factorization is infeasible on commodity hardware.

For \emph{prediction alone} this cost is paid once. For \emph{ARD} it is paid
many times, because evidence maximization is iterative: every step of the
optimizer requires a fresh evaluation of \eqref{eq:logml} and its gradient at a
new $\btheta$, and hence a fresh factorization. A gradient-based optimizer might
take tens to hundreds of steps, and the multi-start strategy of
Section~\ref{sec:landscape} multiplies this by the number of restarts. The
effective cost of fitting an ARD model is therefore the cubic factorization cost
times the number of likelihood evaluations times the number of restarts---a
product that explains why naive ARD does not scale.

A concrete accounting makes the point vivid. Suppose $N = 5000$, a modest size by
contemporary standards, and suppose the optimizer takes $100$ evaluations per
restart with $10$ restarts. Each Cholesky factorization costs on the order of
$N^3/3 \approx 4\times 10^{10}$ floating-point operations; multiplied by
$100\times 10 = 1000$ evaluations, the fit requires on the order of $4\times
10^{13}$ operations, dominated entirely by the factorizations. Doubling $N$ to
$10{,}000$ multiplies this by eight. The storage, at $N^2$ double-precision
numbers, is $200$\,MB at $N=5000$ and $800$\,MB at $N=10{,}000$, approaching the
limits of comfortable in-memory computation. These figures, rough as they are,
explain both why exact ARD is perfectly comfortable for the small-to-moderate
problems that dominate practice and why it hits a wall well before the data sizes
now common in machine learning---and hence why the approximations of
Section~\ref{sec:scalable} are not a luxury but a necessity above a few thousand
points.

\subsection{Gradients and their cost}
\label{sec:gradients}

The one piece of good news in the exact computation is that the gradient of the
marginal likelihood is cheap \emph{given} the factorization. The gradient itself
was derived in Section~\ref{sec:gradient}: equation \eqref{eq:grad} expresses
$\partial_{\theta_m}\log p(\by\mid X,\btheta)$ as a single trace involving
$\balpha\balpha^\top - \Ky^{-1}$ and the kernel derivative $\partial \Ky/\partial
\theta_m$, with the SE-ARD lengthscale derivative \eqref{eq:dkdl} available in
closed form. What concerns us here is its cost. Once the Cholesky factor $L$ and
the vector $\balpha = \Ky^{-1}\by$ are in hand---both by-products of the single
factorization needed to evaluate the likelihood---each of the $D+2$ partial
derivatives can be formed at $O(N^2)$, or $O(N^2 D)$ in total, which is dominated
by the $O(N^3)$ factorization itself. Analytic gradients of this kind are what
make gradient-based evidence maximization practical, and they are built into every
GP library; the binding constraint is the factorization that precedes them, not
the gradient. Modern automatic-differentiation frameworks compute the same
gradient without hand-coding \eqref{eq:dkdl}, and matrix-free implementations
(Section~\ref{sec:scalable}) replace the explicit $\Ky^{-1}$ in \eqref{eq:grad}
with stochastic trace estimates, but the accounting is unchanged: the cost of the
gradient is the cost of solving with $\Ky$.

\subsection{Ill-conditioning and the nugget}
\label{sec:conditioning}

A second exact-computation difficulty is numerical. The matrix $\Kff$ becomes
ill-conditioned---nearly singular---when inputs are close together or when
lengthscales are large, both of which occur routinely during ARD optimization, as
irrelevant inputs push their lengthscales upward. An ill-conditioned $\Kff$
breaks the Cholesky factorization. The standard remedy is to add a small
\emph{jitter} or \emph{nugget} $\tau^2 I$ to the diagonal, which in the regression
model \eqref{eq:model} is naturally absorbed into the noise variance
$\sigma_n^2$. The nugget stabilizes the factorization at the cost of a slight bias
toward smoother fits, and its size must be chosen large enough for stability but
small enough not to distort the model---a minor but real piece of practical
craft that every GP practitioner learns.

\subsection{The dimensional burden of ARD specifically}
\label{sec:dimburden}

The difficulties above afflict all GPs. ARD adds one of its own: the
hyperparameter vector has length $D+2$, growing with the input dimension. A
larger hyperparameter space means a harder optimization---more directions in
which to get stuck, a stronger case for multiple restarts, and a greater chance
of the evidence overfitting documented by \citet{qi2004}. There is also a
statistical dimension to the burden: as Section~\ref{sec:disadvantages} noted,
the reliability of lengthscale thresholding degrades as $D$ grows, because the
relevances of irrelevant inputs spread out and crowd the threshold. High
dimension is thus doubly hard for ARD, computationally and statistically, and it
is the regime in which the alternatives of Section~\ref{sec:compare}---especially
strongly regularized fully Bayesian priors \citep{eriksson2021} and scalable
spike-and-slab \citep{dance2022}---most clearly earn their additional cost.

\subsection{Scalable approximations}
\label{sec:scalable}

The cubic bottleneck has driven two decades of work on approximate GPs, surveyed
comprehensively by \citet{liu2020}. The good news for our purposes is that nearly
all of these approximations are compatible with ARD kernels: one can have
scalability and per-dimension relevance at the same time. We sketch the main
families.

\paragraph{Inducing-point and sparse variational methods.} The dominant approach
summarizes the $N$ training points by $M \ll N$ \emph{inducing points}, reducing
the cost from $O(N^3)$ to $O(NM^2)$. Early versions \citep{snelson2006}, unified
by \citet{quinonero2005}, modified the model; the variational treatment of
\citet{titsias2009} instead approximates the posterior while leaving the model
intact, choosing the inducing points by optimizing a bound on the marginal
likelihood. Introducing inducing variables $\mathbf{u}$ (the latent function at
the inducing inputs) and a variational distribution $q(\mathbf{u})$, this bound is
the evidence lower bound (ELBO)
\begin{equation}
\label{eq:elbo}
\begin{split}
\mathcal{F}
&= \sum_{i=1}^N \E_{q(f_i)}\!\big[\log p(y_i \mid f_i)\big]
\;-\; \mathrm{KL}\!\big(q(\mathbf{u}) \,\|\, p(\mathbf{u})\big)\\
&\le\; \log p(\by \mid X,\btheta),
\end{split}
\end{equation}
whose first term is a sum over data points (hence amenable to mini-batching) and
whose second term regularizes the inducing distribution toward the prior.
Crucially for us, the ARD lengthscales enter $\mathcal{F}$ through the kernel and
are optimized \emph{jointly} with the inducing-point locations and the variational
parameters by maximizing \eqref{eq:elbo}, rather than by maximizing the exact
marginal likelihood. \citet{hensman2013} cast this bound in a form amenable to
stochastic gradient descent, yielding the stochastic variational GP (SVGP) that
scales to millions of points by processing mini-batches; the same construction
extends to non-Gaussian likelihoods \citep{hensman2015}, so SVGP with an ARD
kernel retains relevance learning for classification and count data as well as
regression.

\paragraph{Structured and low-rank methods.} When inputs lie on or near a grid,
structured kernel interpolation (KISS-GP) of \citet{wilson2015} exploits Kronecker
and Toeplitz structure for near-linear scaling. A practical caveat for ARD is
that some structured backends assume additive or low-dimensional structure and do
not directly accommodate a full $D$-dimensional product ARD kernel in high
dimension; this is an implementation limitation rather than a fundamental one, but
it is worth knowing when choosing software. Nystr\"om and other low-rank
factorizations \citep{liu2020} provide further options, and random-feature
approximations---which replace the kernel by an explicit finite-dimensional
feature map---offer another route to near-linear cost for kernel and GP regression
\citep{zheng2024}. Quite apart from
algorithmic approximation, modern hardware-aware implementations change the
practical constant: the blackbox matrix--matrix (BBMM) inference of
\citet{gardner2018} recasts all training and prediction quantities in terms of
batched matrix multiplications solved by conjugate gradients with a pivoted-
Cholesky preconditioner, reducing exact GP inference to $O(N^2)$ and exploiting
GPU parallelism, so that ARD models which were once limited by wall-clock time can
now be fitted at substantially larger $N$ without any change to the model.

\paragraph{Nearest-neighbor methods for large $N$ and large $p$.} A complementary
sparse approach, originating in spatial statistics, replaces the full conditional
structure with a directed graph in which each observation conditions only on a
small set of nearest neighbors. The nearest-neighbor GP (NNGP) of
\citet{datta2016} yields a valid sparse precision matrix and near-linear cost, and
it has proved a natural vehicle for variable selection in the large-$p$ regime that
ARD finds hardest. \citet{posch2025} condition the NNGP mean and covariance on a
random subset of variable indices and place reference priors on the remaining
parameters, performing selection by Markov chain Monte Carlo while keeping the
covariance numerically robust even when candidate variables vastly outnumber
observations. In genomics, the NNGP-based \texttt{nnSVG} method of
\citet{weber2023} fits a per-gene lengthscale and scales linearly in the number of
spatial locations, identifying spatially variable genes at a scale exact GPs
cannot reach. These methods illustrate that the large-$N$ and large-$p$ difficulties
of ARD, long treated as fundamental, are increasingly addressed by the same sparse-
covariance machinery.

\paragraph{Vecchia approximations.} Originating in spatial statistics
\citep{vecchia1988}, the Vecchia approximation factorizes the joint density into a
product of conditionals, each conditioning on a small number of near neighbors,
yielding a sparse Cholesky factor and near-linear cost. \citet{cao2022} built a
Vecchia-based method (VGPR) for simultaneous GP regression and variable selection
that operates at a scale far beyond exact GPs. It optimizes a \emph{penalized}
Vecchia log-likelihood, traversing a regularization path from strong to weak
penalization while sequentially adding promising covariates by the gradient and
deselecting irrelevant ones through an iterative adaptive bridge penalty, with a
Vecchia-specific mini-batch subsampling scheme that yields unbiased gradient
estimates. The result is a method that, on a problem with $N = 10^6$ observations
and $D = 10^3$ candidate inputs, runs in well under an hour, comparable to the
time taken by the lasso and by regression trees on the same data. This line of
work shows that the cubic bottleneck, long the binding constraint on GP variable
selection, is no longer fundamental---ARD-style relevance learning can now be
carried out at genuinely large scale, and the bridge penalty makes explicit the
link to the penalized methods of Section~\ref{sec:penalized}. Related work
attacks the same large-scale covariance-estimation problem from the spectral side:
\citet{mosammam2025} estimates space-time covariance functions for large
Gaussian-random-field datasets by frequency-domain composite likelihood, avoiding
direct factorization of the full covariance matrix altogether.

\paragraph{The upshot.} The practical recommendation is dimensional. For small to
moderate $N$ (up to a few thousand) exact GP computation with ARD is fine and is
what the libraries do by default. For large $N$ one couples ARD with a scalable
approximation---SVGP for general inputs, Vecchia or NNGP for spatial or
moderate-dimensional problems---and pays a constant-factor overhead rather than a
cubic one; for large $p$, nearest-neighbor and spike-and-slab variational
constructions keep both the computation and the selection tractable. The choice of
approximation interacts with the input dimension and the data geometry, but in all
cases the ARD relevances remain available as a by-product of the (approximate)
marginal-likelihood or variational optimization.

\section{Evaluation Frameworks and Application Domains}
\label{sec:applications}

Before surveying where ARD is used, we address a question the preceding sections
have deferred: once a method has produced a selection, how is its quality judged?
A survey that recommends methods owes the reader a vocabulary for comparing them,
and the criteria differ sharply depending on whether selection or prediction is
the goal.

\subsection{Evaluating variable selection}
\label{sec:evaluation}

When the active set is known---as in the synthetic illustrations of
Sections~\ref{sec:motivation} and~\ref{sec:disadvantages}, and in simulation
studies generally---selection quality is measured by the familiar confusion-matrix
quantities. Writing $S$ for the selected set and $S_0$ for the true active set,
\emph{precision} (the fraction of selected inputs that are truly relevant,
$|S \cap S_0|/|S|$) and \emph{recall} or power (the fraction of relevant inputs
recovered, $|S \cap S_0|/|S_0|$) summarize the two ways selection can err, and the
\emph{false discovery rate}, the expected value of $|S \setminus S_0|/|S|$, is the
quantity that the knockoff methods of Section~\ref{sec:thresholding} control by
construction. Because ARD produces a ranking rather than a set
(Section~\ref{sec:effectivesparsity}), it is naturally evaluated by sweeping the
threshold and tracing a receiver-operating-characteristic or precision--recall
curve, which separates the quality of the \emph{ranking} from the quality of the
\emph{threshold}---a useful distinction, since ARD's ranking is often good even
where its threshold is unreliable. How reliably any method recovers the true
support depends sharply on the design, and correlation among the candidate inputs
in particular: \citet{kumar2024} characterize numerically how support recovery in
sparse regression degrades as the predictors become correlated, a cautionary
finding that applies equally to relevance-based selection, whose lengthscales are
hardest to separate precisely when inputs are collinear.

For Bayesian methods that return posterior inclusion probabilities, a further and
more stringent criterion is \emph{calibration}: among inputs assigned inclusion
probability near $p$, a well-calibrated method should find that a fraction near
$p$ are truly relevant. Calibration is what distinguishes an honest probabilistic
selection from a mere score, and it is a principal advantage claimed for the
spike-and-slab and fully Bayesian methods of Section~\ref{sec:compare} over
point-estimate ARD.

When prediction rather than selection is the goal, the relevant metrics are
predictive: out-of-sample mean-squared error or, better for a probabilistic model,
the out-of-sample log predictive density, which rewards calibrated uncertainty and
not merely accurate point predictions. The two families of metrics can disagree---a
model can predict well while selecting poorly, or recover the active set while
predicting no better than a denser competitor---and which disagreement one
tolerates depends on the application, a tension that the projection-predictive
methods of Section~\ref{sec:sensitivitygroup} were designed to manage by selecting
explicitly for predictive performance. The practical recommendation is to report
both kinds of metric and to be explicit about which one the selection is meant to
optimize.

\subsection{Application domains}
\label{sec:domains}

ARD-based variable selection is used across the quantitative sciences wherever a
Gaussian process is fitted and the relevance of inputs is in question. We
highlight the domains in which it has been most consequential.

\paragraph{Computer experiments and emulation.} The historical home of GP
variable selection is the design and analysis of computer experiments
\citep{sacks1989,santner2018}, where an expensive deterministic simulator is
emulated by a GP and the analyst wishes to screen a long list of inputs for the
few that drive the output. The setting is almost tailor-made for ARD: the
simulator is typically smooth, deterministic (so the noise is pure nugget for
numerical stability), and expensive enough that only a modest number of runs are
affordable, which keeps $N$ in the range where exact ARD is comfortable while
making automatic screening genuinely valuable. \citet{welch1992} pioneered a
constrained-maximum-likelihood procedure---closely related to ARD---that adds
lengthscale parameters one at a time, building up the set of relevant inputs
greedily, and this forward-screening philosophy remains standard in emulation
toolkits to this day. \citet{linkletter2006} brought the inert-reference device to
this setting, giving a principled way to decide when an estimated relevance is
distinguishable from noise, and the spike-and-slab framework of
\citet{savitsky2011} was developed with computer-experiment and related
applications in view. The choice of input design feeds directly into how well
relevances can be estimated, and the construction of space-filling designs whose
quality is assessed through the Kriging (GP) emulator remains an active topic
\citep{lai2024}. Emulation is the setting in which ARD's combination of
nonlinearity, uncertainty, and automatic screening is most naturally suited, and
it is no accident that much of the methodology surveyed here was first developed
or first applied there.

\paragraph{Spatial statistics and geostatistics.} In kriging, the lengthscales of
the covariance are the spatial ranges, and learning them is part of standard
practice \citep{stein1999}. ARD-type anisotropic ranges distinguish directions of
rapid and slow spatial variation, and penalized and tapered methods
\citep{chu2011} bring explicit selection to the spatial-regression setting. The
identifiability subtleties of \citet{zhang2004} were first appreciated here, and
they temper the interpretation of estimated ranges as relevances.

\paragraph{Genomics and the life sciences.} High-dimensional biological data,
where the number of candidate predictors vastly exceeds the sample size, were an
early target of ARD-style sparse Bayesian methods and remain an active one.
\citet{li2002} applied Bayesian ARD algorithms to classify gene-expression data,
and the nonparametric spike-and-slab GP of \citet{savitsky2011} was motivated in
part by such problems. More recent work pushes the same ideas to modern
genomic scales: Gaussian process latent variable models with relevance-determining
priors are used to separate biological from technical variation in single-cell
transcriptomics \citep{lalchand2022}, the NNGP-based methods of
Section~\ref{sec:scalable} identify spatially variable genes at the scale of whole
tissue sections \citep{weber2023}, and Gaussian process models with Bayesian
variable selection map function-valued quantitative traits from incomplete
phenotypic data \citep{vanhatalo2019}. The appeal throughout is the ability to
find a small relevant subset among thousands of candidates while modeling nonlinear
effects and quantifying uncertainty.

\paragraph{Engineering design, sensitivity analysis, and Bayesian optimization.}
In engineering and the physical sciences, GP surrogates support global sensitivity
analysis---apportioning output variation among inputs---and ARD lengthscales serve
as a quick screening tool, often combined with active subspaces
\citep{constantine2014} when the relevant structure is a rotation of the inputs.
In Bayesian optimization of expensive black-box functions, relevance learning
focuses the search on the inputs that matter; the SAAS prior of
\citet{eriksson2021} was developed precisely to make high-dimensional Bayesian
optimization tractable by aggressively selecting a sparse axis-aligned subspace,
and it has become a standard tool for optimization problems in tens to hundreds of
dimensions where an unstructured GP would be hopeless. The common thread in these
engineering applications is that selection and prediction serve a downstream
decision---which design to build, which experiment to run next, which parameters
to tune---so that the value of relevance learning is measured not in isolation but
by its effect on the quality of those decisions, a perspective that favors methods
like ARD that integrate selection seamlessly into a predictive workflow.

\paragraph{A common pattern.} Across these domains a common pattern holds: ARD is
used first because it is immediate, and the more elaborate methods of
Section~\ref{sec:compare} are brought in when the dimension, the need for exact
selection, or the demand for calibrated uncertainty exceeds what plain lengthscale
thresholding can deliver. The applications also illustrate why no single method has
displaced the others---computer experiments favor ARD and its screening relatives,
high-dimensional optimization favors strongly regularized Bayesian priors, spatial
problems favor tapered and Vecchia methods, and interaction-focused analyses favor
additive models---so that the diversity of the methodological landscape mirrors the
diversity of the problems it serves.

\section{Discussion and Open Problems}
\label{sec:discussion}

Automatic relevance determination occupies a curious and instructive position in
statistical methodology. It is, by the standards of the alternatives surveyed in
Section~\ref{sec:compare}, the least principled option: it produces only effective
sparsity, offers no calibrated inclusion probabilities, propagates no
hyperparameter uncertainty in its standard point-estimate form, can misjudge
relevance when linear and nonlinear effects are mixed, and comes with no general
selection-consistency guarantee. And yet it is, by a wide margin, the most widely
used. This survey has tried to explain both halves of that paradox.

The explanation for ARD's success is that it solves a real problem at almost no
cost. The mechanism is sound: the Bayesian Occam's razor embodied in the marginal
likelihood (Section~\ref{sec:evidence}) genuinely does drive irrelevant inputs
toward large lengthscales, and in the low-to-moderate-dimensional regime where
most GP modeling happens, the resulting relevance ranking is usually correct. The
cost is negligible: ARD is the default anisotropic kernel, fitted by the same
optimization one would run anyway, available as a single flag in every library.
For the working statistician who wants a quick, interpretable, uncertainty-aware
sense of which inputs matter, ARD is hard to beat, and its dominance is rational
rather than merely habitual.

The explanation for ARD's limitations is equally clear, and it sharpens into
practical guidance. The method's failure modes are concentrated in identifiable
regimes---high input dimension, the need for exact zeros or calibrated
probabilities, mixed effect types, and demands for formal guarantees---and in each
of those regimes there is a well-developed alternative that addresses the specific
deficiency. Scalable spike-and-slab GPs \citep{dance2022} supply exact sparsity and
inclusion probabilities at nearly ARD's cost; strongly regularized fully Bayesian
priors such as SAAS \citep{eriksson2021} select reliably in high dimension;
sensitivity and projection methods \citep{piironen2016,paananen2019} measure
relevance by effect on prediction rather than by rate of variation; additive and
compositional kernels \citep{duvenaud2011,duvenaud2013} expose interaction
structure; active subspaces \citep{constantine2014} handle rotated structure; and
penalized kriging \citep{chu2011} brings consistency theory. The mature view of GP
variable selection is therefore not ``ARD versus the rest'' but a decision tree:
start with ARD, and escalate to the alternative matched to whichever of ARD's
limitations the problem actually exposes.

Several developments are narrowing ARD's limitations even within its own
framework. The cubic bottleneck, long the binding computational constraint, has
been substantially relaxed by inducing-point, structured, and Vecchia
approximations (Section~\ref{sec:scalable}), to the point where \citet{cao2022}
perform GP variable selection at a scale of a million observations and a thousand
inputs. The high-dimensional selection problem is being addressed by the marriage
of the continuous lengthscale parameterization with strong sparsity priors and
honest integration \citep{eriksson2021}, which retains ARD's parameterization while
curing its point-estimate pathologies. And the gap between ARD's heuristic
relevances and a principled selection is being closed from the variational side by
methods that bolt inclusion probabilities onto scalable GP inference
\citep{dance2022}.

We close with the open problems we consider most pressing. First, a
selection-consistency theory directly for marginal-likelihood ARD---perhaps under
design or smoothness restrictions that restore the identifiability that
\citet{zhang2004} shows is generically absent---would put the most-used method on
the firm footing its competitors enjoy. One promising avenue is to modify the
estimation objective so that it targets the well-identified microergodic
parameters rather than the confounded individual lengthscales: the jointly robust
priors of \citet{gu2019}, which by construction separate influential from inert
inputs and come with favorable robustness properties for parameter estimation,
suggest that a carefully designed prior or penalty on the marginal likelihood
might recover the identifiability that raw ARD lacks. Second, the
linear-versus-nonlinear misranking of \citet{paananen2019} deserves a constructive
fix: a relevance measure that is as cheap and automatic as a lengthscale but that
tracks contribution to prediction rather than rate of variation. Third, the choice of
thresholding rule (Section~\ref{sec:thresholding}) remains the weakest link in
relevance-based selection, and a default, well-calibrated, theoretically justified
rule---going beyond the useful but ad hoc inert-reference device---would benefit
every practitioner. Progress on any of these would strengthen a method that, for
all its imperfections, has earned its place as the first tool reached for when a
Gaussian process must decide what matters.

Beyond these specific problems, several frontiers will shape the next decade of
GP variable selection. The stationarity assumption that underlies the SE and
Mat\'ern kernels---and hence the lengthscale notion of relevance---fails for
functions with discontinuities or abrupt regime changes, and a recent line of work
on jump and piecewise GP surrogates \citep{park2025, sauer2023} is beginning to
ask what relevance even means across a discontinuity, a question with no settled
answer. The newest of these methods push into exactly the high-dimensional regime
where relevance matters most: the deep jump GP of \citet{xupark2025} composes a
region-specific locally linear projection with a jump GP, learning a low-
dimensional active subspace separately within each piece of a piecewise-continuous,
high-dimensional function---in effect performing localized relevance determination
on either side of a discontinuity. The extension of ARD beyond Gaussian noise,
already practical through scalable variational inference for classification and
point-process likelihoods \citep{hensman2015}, is increasingly important as GPs are
applied to count data, survival outcomes, and the analysis of neural recordings,
where relevance learning must operate inside a non-Gaussian observation model. And
the move from single-output to multi-output and multi-task GPs raises the question
of relevance that is shared across tasks versus specific to one, a structure that
plain ARD does not express. In each of these directions the core ARD idea---a
learned, per-input notion of relevance read off the covariance---must be rethought
rather than merely ported, and that rethinking is where much of the methodological
action now lies.

Stepping back, the trajectory of the field suggests a synthesis rather than a
victor. The continuous lengthscale parameterization that defines ARD, the discrete
inclusion indicators of spike-and-slab, the heavy-tailed shrinkage of SAAS, the
structured decompositions of additive models, and the prediction-centered criteria
of the sensitivity methods are not so much competing answers as complementary
expressions of a single underlying question---how to let a flexible nonparametric
model concentrate on the few inputs that matter---and the most effective modern
methods increasingly combine them, placing strong sparsity priors on lengthscales,
bolting inclusion probabilities onto scalable variational inference, or wrapping
ARD components inside additive structures. ARD's lasting contribution may be less
the specific device of per-input lengthscales than the demonstration that
relevance can be \emph{learned}, automatically and as a by-product of fitting,
rather than imposed in advance; that idea, which MacKay and Neal introduced for
neural networks and which the GP literature refined into the methods surveyed
here, is now so thoroughly absorbed into nonparametric practice that its origin in
ARD is easy to forget. A survey is a good place to remember it.

\section*{Acknowledgments} 
The authors would like to thank the editor and the reviewers for their valuable comments and suggestions. 

\section*{Author contributions} 
CRediT: Jia Cai: Conceptualization, Data curation, Formal analysis, Methodology, Writing – original draft,Writing – review \& editing.

\section*{Disclosure statement} 
The author reports there are no competing interests to declare.

\section*{Funding} 
No funding is to report.

\bibliographystyle{apalike}
\bibliography{bibliography}

\end{document}